\documentclass[wcp]{jmlr}

\usepackage{longtable}

\usepackage{booktabs}

\title[Fast Dynamic Convolutional Neural Networks]{Fast Dynamic Convolutional Neural Networks for Visual Tracking}

  \author{\Name{Zhiyan Cui} \Email{cuizhiyan@stu.xjtu.edu.cn}\\
  \addr State Key Laboratory for Manufacturing System Engineering, System Engineering Institute, Xi'an Jiaotong University, Xi'an, Shaanxi, China
  \AND
  \Name{Na Lu\textsuperscript{*}} \Email{lvna2009@mail.xjtu.edu.cn}\\
  \addr State Key Laboratory for Manufacturing System Engineering, System Engineering Institute, Xi'an Jiaotong University, Xi'an, Shaanxi, China\\
  Beijing Advanced Innovation Center for Intelligent Robots and Systems, Beijing, China
 }

\begin{document}
\maketitle
\begin{abstract}
Most of the existing tracking methods based on CNN(convolutional neural networks) are too slow for real-time application despite the excellent tracking precision compared with the traditional ones. In this paper, a fast dynamic visual tracking algorithm combining CNN based MDNet(Multi-Domain Network) and RoIAlign was developed. The major problem of MDNet also lies in the time efficiency. Considering the computational complexity of MDNet is mainly caused by the large amount of convolution operations and fine-tuning of the network during tracking, a RoIPool layer which could conduct the convolution over the whole image instead of each RoI is added to accelerate the convolution and a new strategy of fine-tuning the fully-connected layers is used to accelerate the update. With RoIPool employed, the computation speed has been increased but the tracking precision has dropped simultaneously. RoIPool could lose some positioning precision because it can not handle locations represented by floating numbers. So RoIAlign, instead of RoIPool, which can process floating numbers of locations by bilinear interpolation has been added to the network. The results show the target localization precision has been improved and it hardly increases the computational cost. These strategies can accelerate the processing and make it $7\times$ faster than MDNet with very low impact on precision and it can run at around 7 fps. The proposed algorithm has been evaluated on two benchmarks: OTB100 and VOT2016, on which high precision and speed have been obtained. The influence of the network structure and training data are also discussed with experiments.
\end{abstract}

\section{Introduction}
Visual tracking is one of the fundamental problems in computer vision which aims at estimating the position of a predefined target in an image sequence, with only its initial state given. Nowadays, CNN has achieved great success in computer vision, such as object classification~\cite{22,30}, object detection~\cite{10,9,28}, semantic segmentation~\cite{10,14,11} and so on. However, it is difficult for CNN to play a great role in visual tracking. CNN usually consists of a large amount of parameters and needs big dataset to train the network to avoid over-fitting. So far, there are some novel studies~\cite{23,5,27,34} which combine CNN and traditional trackers to achieve the start of the art. Some of them just use CNNs which are trained on ImageNet or other large datasets to extract features.\\
On one hand, tracking seems much easier than detection because it only needs to conduct a binary classification between the target and the background. On the other hand, it is difficult to train the network because of the diversity of objects that we might track. An object may be the target in one video but the background in another. And there is no such amount of data for tracking to train a deep network.\\
MDNet~\cite{26} is a novel solution for tracking problem with detection method. It is trained by learning the shared representations of targets from multiple video sequences, where each video is regarded as a separate domain. There are several branches in the last layer of the network for binary classification and each of them is corresponding to a video sequence. The preceding layers are in charge of capturing common representations of targets. While tracking, the last layer is removed and a new single branch is added to the network to compute the scores in the test sequences. Parameters in the fully-connected layers are fine-tuned online during tracking. Both long and short updating strategies are conducted for robustness and adaptability, respectively. Hard negative mining~\cite{32} technique is involved in the learning procedure. MDNet won the first place in VOT2015 competition. Although the effect is amazing, MDNet runs at very low speed which is 1 fps on GPU. 
In this paper, we propose a similar but much faster network for object tracking. It consists of 3 convolution layers and 3 fully-connected layers with a RoIAlign layer~\cite{14} between them. The convolution is operated on the whole image instead of each RoI, and a RoIAlign layer is used to get fixed-size features which can enhance the calculation speed. When we compare the results of RoIAlign and RoIPool~\cite{9}, it shows that RoIAlign could obtain higher precision. Instead of fine-tuning the network with fixed number of iterations, we use a threshold to adjust the numbers dynamically, which will reduce the iterations and time of fine-tuning. Similar training and multi-domain strategy as in MDNet have been adopted. Given $k$ videos to train the network, the last fully-connected layer should have $k$ branches. Each video is corresponding to one branch. So the last layer is domain-specific layers and others are shared layers. In this way, the shared information of the tracked objects is captured.\\
The $k$ branches of last layer are deleted and a new one is added to the network before tracking. The parameters in last layer are initialized randomly. The positive and negative samples which are extracted from the first frame of the sequence are used to fine-tune the network. Only parameters in the fully-connected layers will be updated. Bounding boxes in the current frame around the previous target position are sent to the network for discrimination. The sample with the highest score will be obtained as the target in this frame. The fully-connected layers will be updated with samples from the previous several frames when scores of all the samples are less than a threshold. By this dynamic method, the network can learn the new features of the objects.\\
The proposed algorithm consists of multi-domain learning and dynamic network tracking. The main contributions are shown below:\\
\begin{itemize}
\item A small neural network with RoIAlign layer is created to accelerate the convolution. 
\item A new strategy of fine-tuning the fully-connected layers is used to reduce the iterations during tracking. About 7 fps on GPU could be reached which makes it $7\times$ faster than MDNet with only a small decrease in precision.
\item The performance of different structures of the network has been compared. The result shows that RoIAlign outperforms RoIPool which is usually used in detection and classification. And neither increasing convolution layers nor decreasing fully-connected layers has positive effects on the accuracy. 
\item 	Evaluations of the proposed tracker have been performed on two public benchmarks: object tracking benchmark~\cite{36} and VOT2016~\cite{44}. The results show that a higher precision and faster tracking speed than most of the existing trackers based on CNN have been obtained.	
\end{itemize}
\section{Related Work}
There have been several surveys~\cite{39,38} of visual tracking during these years. Most of the state of the art methods are based on correlation filter or deep learning. Some trackers have been combined with each other to improve the tracking performance.\\
Correlation filter (CF) has attracted scientists’ attention for several years. It can run at high speed because of the application of Fourier transformation. The calculation is transformed from spatial domain to Fourier domain to speed up the operation. MOSSE~\cite{4} was the first algorithm which uses CF to track objects. CSK~\cite{16} is another algorithm based on CF. It uses circular shift to produce dense samples by making use of more features of the image. And then, there are KCF~\cite{17} and DCF~\cite{17}. Both of them can use multi-channel features of images. DSST~\cite{6} uses translation filter and scale filter to deal with changes in location and scale. Because of Fourier transformation, most of these trackers can reach a very high speed.\\ 
Nowadays, deep learning has been widely used in computer vision, such as classification, detection and semantic segmentation. Many trackers based on deep learning have been designed to get a better precision. Some of them combine deep learning with CF. DLT~\cite{37} is the first tracker to use deep learning. The structure is based on particle filter and it uses SDAE to extract features. HCF~\cite{23} uses the features in different layers from a neural network and then applies CF to conduct tracking. MDNet~\cite{26} is a novel tracker based on CNN which uses detection method to realize tracking. It samples bounding box around the target location in the next frame and sends them to the network to find the one with the highest score, which is the target in the next frame. And the network will be fine-tuned when it does not work well any longer. TCNN~\cite{25} is a tracker based on CNN and a tree structure. It builds a tree to evaluate the reliability of the model. GOTURN~\cite{15} is the fastest tracker based on CNN and it can achieve 165 fps. It uses the deep neural network to regress the bounding box in the next frame. But its target localization precision is poor. In a word, the high precision of a tracker based on CNN is obtained at the sacrifice of speed.
\section{Fast Dynamic Convolutional Neural Networks}
This section describes the structure of the network and how to train and test the proposed network. The convolution is operated on the whole image and RoIAlign is used to obtain fix-sized features. This strategy can speed up the convolution computation and maintain the localization accuracy in the meanwhile. Different branches have been added to the last layer in correspondence to different video sequences, and the preceding layers are shared layers to get common representations of targets through training. During the testing stage, the last layer is removed and a new one with only one branch is added to the network. Parameters in the fully-connected layers will be updated in a new strategy which will reduce the ineffective updates during tracking to adapt to the currently tracked targets.
\subsection{Convolution on the Whole Image}
In MDNet, there are 256 RoIs extracted from each frame and all of them are scaled to a fixed size. Then they will be sent to the convolutional layers and get fixed-size features which can then be delivered to the fully-connected layers. But it is a waste of time to do so. It contains a lot of redundancy calculation and the convolution operation has to repeat for many times which will definitely increase the time of process.\\
To speed up the convolution, a strategy of conducting convolution on the whole image is adopted and it only needs to do the convolutional operation for once. At the end of the convolution layers, a RoIPool or RoIAlign is used to extracted fixed-size features corresponding to the RoIs from the last convolution layer. The experiments show that RoIAlign is better than RoIPool in precision which we will discuss the details in section 4.
\subsection{Network Architecture}
\begin{figure}[htp]
	\begin{center}
		\includegraphics[width=0.9\textwidth]{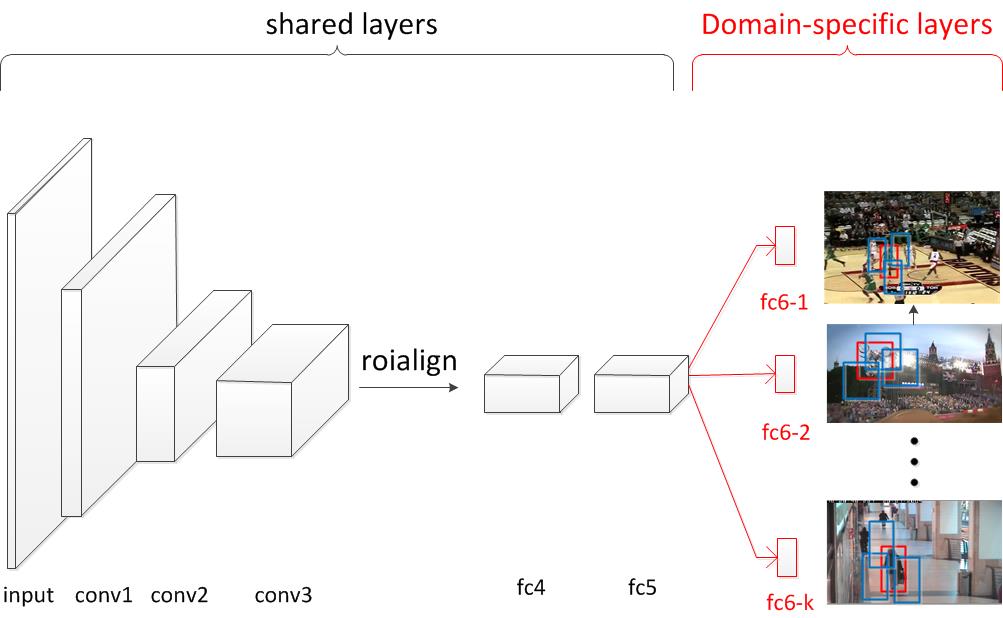}
		\caption{The architecture of fast dynamic convolutional neural network. Red and blue bounding boxes denote the positive and negative samples, respectively.}\label{fig:network}
	\end{center}
\end{figure}

\begin{figure}[htp]
	\begin{center}
		\includegraphics[width=0.9\textwidth]{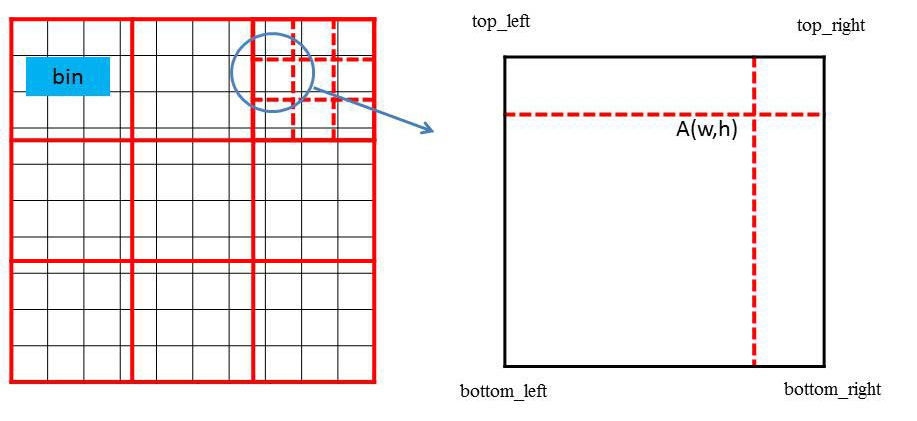}
		\caption{The schematic diagram of RoIAlign. Suppose there is a 10x10 pixels RoI (in black) and we want to get a 3x3 feature (in red). 2x2 points are selected in each bin. Because the locations of the points are float types, we use four integer points nearby and bilinear interpolation to get the value of this point. And the biggest one is used to represent this bin (max RoIAlign).}\label{fig:align}
	\end{center}
\end{figure}

The architecture of our network is illustrated in Fig.~\ref{fig:network}. The first three convolution layers are borrowed from the VGG-M~\cite{45} network. The size of the filters in the first convolution layer is $7\times7\times3\times96$, which is followed by a Relu, a normalization~\cite{19} and a pooling layer. The size of the filters in the second convolution layer is $5\times5\times96\times256$, also followed by a Relu, a normalization and a pooling layer. The size in the third convolution layer is $3\times3\times256\times512$, followed by a Relu layer. Then there is a RoIAlign layer to extract fixed size features (here we set it to $3\times3\times512$) from the convolution layers. Next, there are two fully-connected layers with dropout (the rate is set to 0.5). The last fully-connected layer is just like the one in MDNet. Suppose we use $k$ videos to train the network, the network will have $k$ branches in the last layer and each of them uses two labels to represent the target and the background. The last layer is domain-specific layer and the others are shared layers. That means, when the network is trained with the $i\textsuperscript{th}(i=1,2,…,k)$ video, we just update the parameters in $i\textsuperscript{th}$ branch of the last fully-connected layer and the shared layers.
\\
Most of the neural networks for classification and detection use Pool or RoIPool layer after the convolution layers. RoIPool can get fixed-size (e.g., $3\times3$) features from each RoI, which can then be sent to the following fully-connected layers. However, there is one major issue with RoIPool. After several convolution operations, the size and position of the RoIs might be float numbers, and we need to divide the RoIs into fixed-size regions (e.g., $3\times3$). The RoIPool rounds the float numbers to the nearest integers to fulfill the pooling. The localization precision may get lost in this operation.\\
RoIAlign~\cite{14} is another way to get fixed-size(e.g., $3\times3$) features from each RoI. It can work better than RoIPool in theory. Meanwhile, RoIAlign does not spend much more time than RoIPool. The performance comparison of RoIPool and RoIAlign could be found in section 4. Different from RoIPool, RoIAlign keeps the float numbers in the operation. At the last step of RoIAlign, we use bilinear interpolation~\cite{46} to calculate $n$ points in each bin and use the largest one to represent this bin(max-RoIAlign). \\
Fig.~\ref{fig:align} shows an example of RoIAlign. The point $A(w,h)$ is what we need and its coordinates are float. The four points:$top\_left(x_{left},y_{top})$,$top\_right(x_{right},y_{top})$,$bottom\_left(x_{left},y_{bottom})$ and $bottom\_right(x_{right},y_{bottom})$ around it are the nearest integer points. The value of $A$ could be calculated via the following equations:\\
\hspace*{2.4cm}$V_1=(1-(w-x_{left}))\times(1-(h-y_{top}))\times Value_{top\_left}$\\
\hspace*{2.4cm}$V_2=(1-(x_{right}-w))\times(1-(h-y_{top}))\times Value_{top\_right}$\\
\hspace*{2.4cm}$V_3=(1-(w-x_{left}))\times(1-(y_{bottom}-h))\times Value_{bottom\_left}$\\
\hspace*{2.4cm}$V_4=(1-(x_{right}-w))\times(1-(y_{bottom-h}))\times Value_{bottom\_right}$\\
\hspace*{2.4cm}$Value_A=V_1+V_2+V_3+V_4$
\\
When it comes to back propagation, just like RoIPool, we just use the points which contribute to the network and propagate their errors. Suppose that point $A$ is contributed to the network and its gradient is represented by $Value_{der}$ , and the back propagation is like that:\\
\hspace*{2.4cm}$Value_{top\_left}=(1-(w-\left\lfloor w\right\rfloor))\times (1-(h-\left\lfloor h\right\rfloor))\times Value_{der}$\\
\hspace*{2.4cm}$Value_{top\_right}=(1-(w-\left\lfloor w\right\rfloor))\times (1-(\left\lceil h\right\rceil-h))\times Value_{der}$\\
\hspace*{2.4cm}$Value_{bottom\_left}=(1-(\left\lceil w\right\rceil-w))\times (1-(h-\left\lfloor h\right\rfloor))\times Value_{der}$\\
\hspace*{2.4cm}$Value_{bottom\_right}=(1-(\left\lceil w\right\rceil-w))\times (1-(\left\lceil h\right\rceil-h))\times Value_{der}$\\
For each bin, $2\times2$ points are chosen and we use the largest one to represent this bin. The precision has dropped slightly if $1\times1$ point is used and $3\times3$ points which will increase the calculation indicates that there is almost no difference with $2\times2$. So $2\times2$ points in each bin is an acceptable choice.\\
A smaller network than those which usually used in the detection or classification is constructed in our study. The most serious issue of the deep network solution for image tracking is the low time efficiency. Considering that in the tracking scenario only binary classification is involved, so a small and fast network can satisfy the need of tracking task. Otherwise, with more convolution layers added, the localization information will get diminished over layers which would influence the tracking accuracy. Some other network structures are compared with our network and the results show the superiority of the proposed network. The details could be found in section 4.
\subsection{Training Method}
Softmax cross-entropy loss and SGD are used to update the network. When $i\textsuperscript{th}$ video is used to train the network, $i\textsuperscript{th}$ branch of the last layer and shared layers are working together and their parameters are updated at the same time. Other branches in the last layer are disabled and do not work. It was shown in ~\cite{26} that $k$ branches could get better performance than only one branch in the last layer during training.\\
Samples are extracted around the target from each frame in a video. The samples whose IoU are larger than a threshold $t1$ are used as positive examples, and the ones with IoU less than a threshold $t2$ are used as negative examples. In our experiment, we set $t1$ to 0.7 and $t2$ to 0.5.\\
The three convolution layers have the same structure as in VGG-M network which have been pre-trained on ImageNet. The parameters in the fully-connected layers are initialized randomly which subject to Gaussian distribution. Because the convolution layers have been trained on ImageNet, a smaller learning rate is adopted from the beginning. We set the learning rate to 0.0001 and weight decay to 0.0005. It takes 100 iterations to get the local optimum.
\subsection{Tracking Method}
\begin{figure} 	
	\centering 	
	\includegraphics[width=0.24\linewidth]{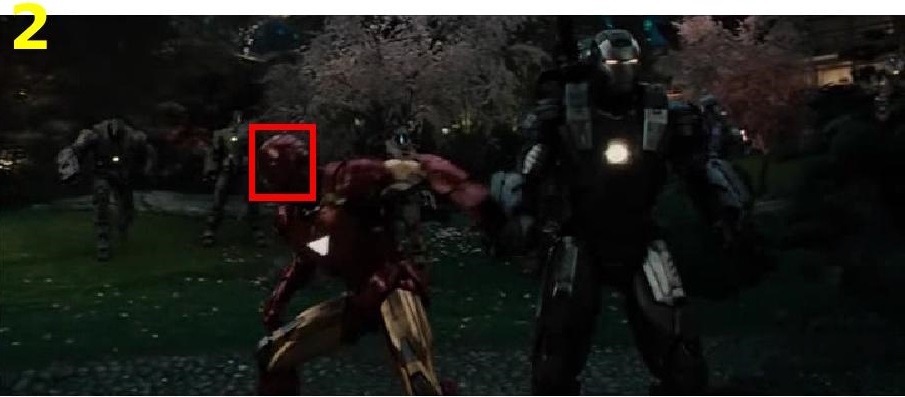}
	\includegraphics[width=0.24\linewidth]{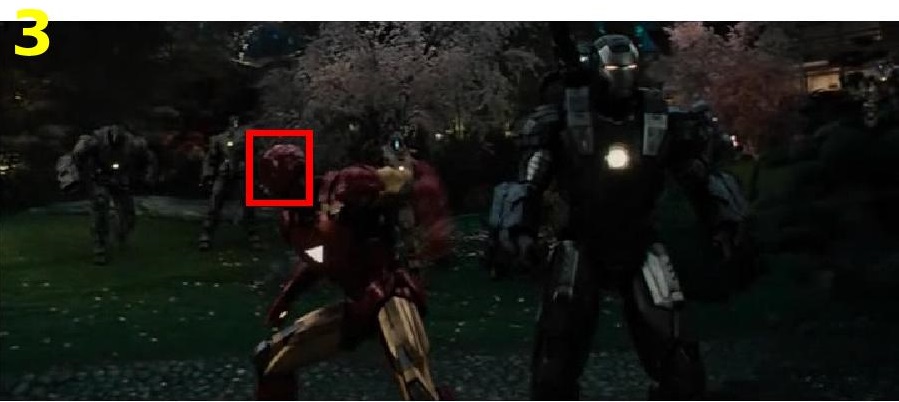}
	\includegraphics[width=0.24\linewidth]{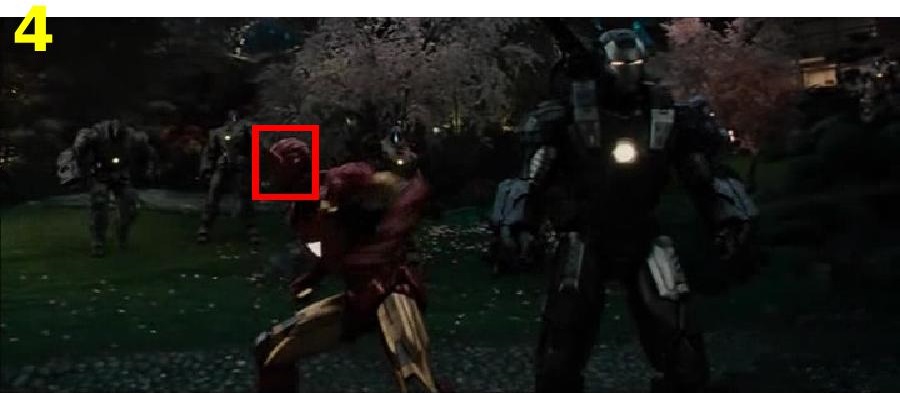}
	\includegraphics[width=0.24\linewidth]{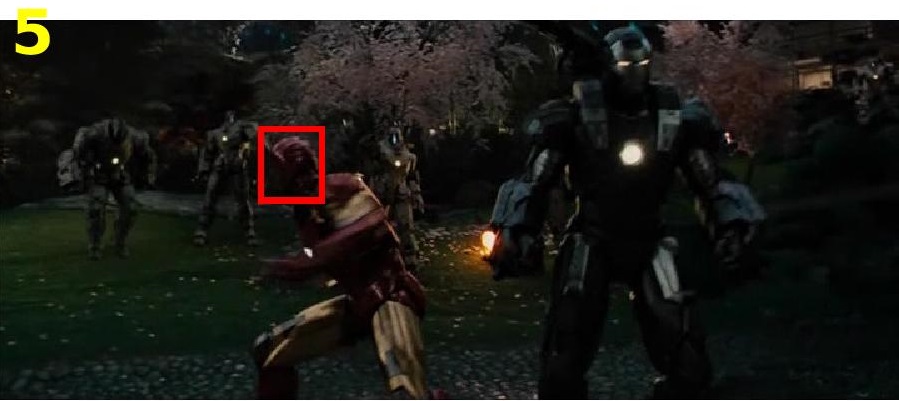}   

	\includegraphics[width=0.24\linewidth]{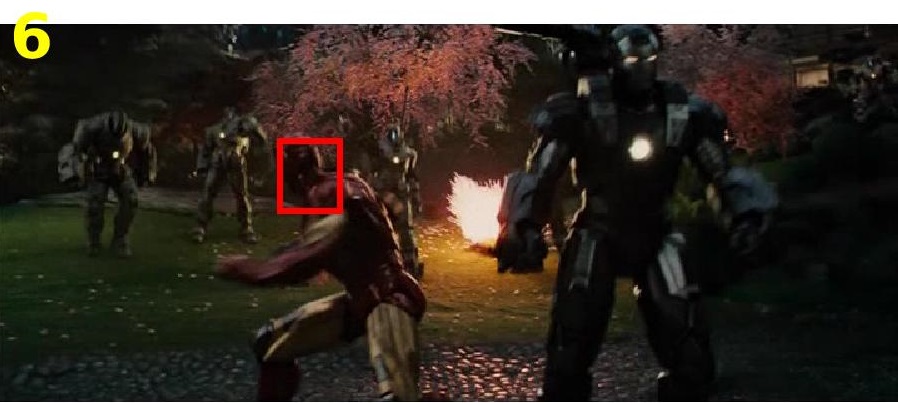}
	\includegraphics[width=0.24\linewidth]{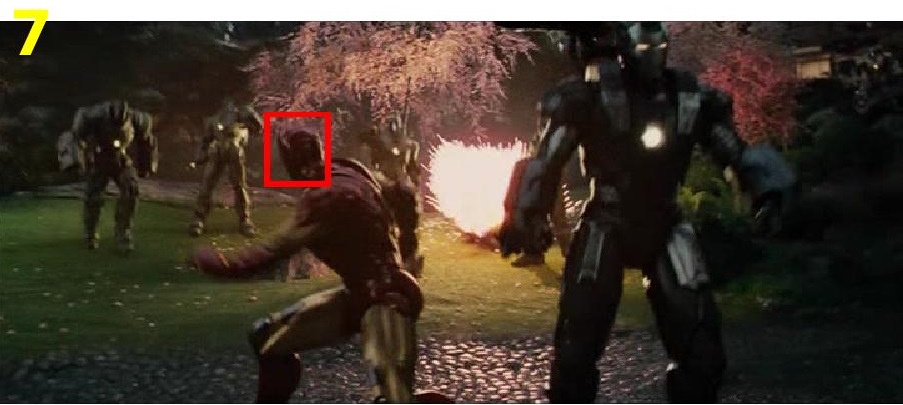}
	\includegraphics[width=0.24\linewidth]{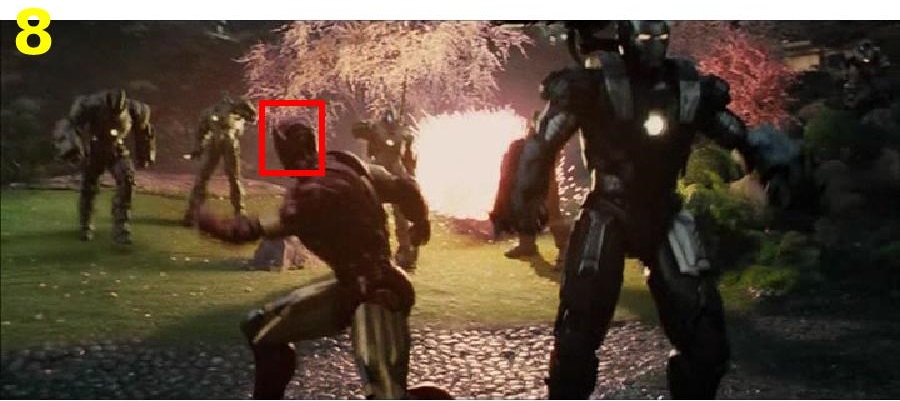}
	\includegraphics[width=0.24\linewidth]{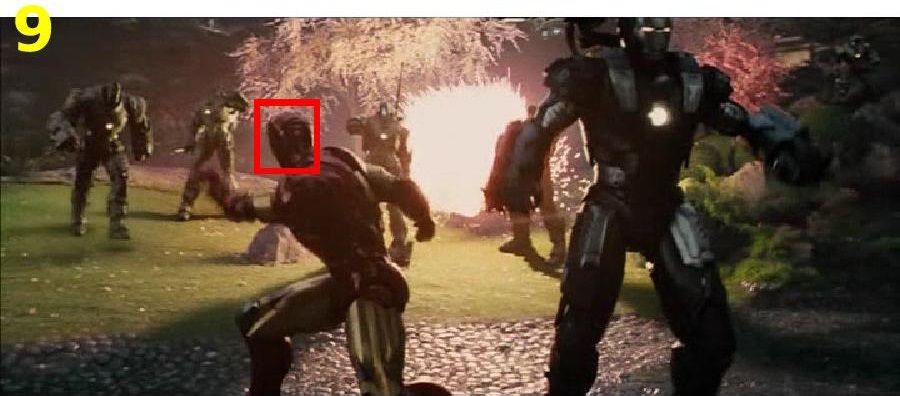}  

	\caption{frame 2-9 in the sequence of Ironman}
	\label{fig:fc_i}	
\end{figure}

After the proposed network has been trained, we delete the last fully-connected layer and add a fully-connected layer with two outputs, representing the target and the background, respectively. The parameters in the last layer are initialized randomly which subject to Gaussian distribution. The three fully-connected layers will be updated while tracking, but the convolution layers will not be changed.\\
In the beginning, the positive and negative examples from the first frame are used to fine-tune the network. Then, we select samples from the current frame near the target and send the samples to the network. The sample with the highest score will be obtained as the target in this frame. If the score is higher than a threshold $m$, both positive and negative samples near the target in the current frame are collected for updating. When the highest score is less than the threshold $m$, we use samples collected from the previous frames to update the fully-connected layers. $m$ is set to 0 in our experiments. \\
In MDNet, the network will be iterated 10 times if the highest score is less than the threshold $m$. But in our experiments, we find that a lot of iterations are useless and spend much time. So we define a loss threshold $l$. The iteration will stop if the loss of the network is less than $l$. It is a very simple strategy, but it is an effective one which will save a large amount of time. We use an example to explain this strategy and its consequence. The loss threshold $l$ is set to 0.01.\\
We use the sequence of Ironman in OTB100. When the score is less than $m$, which is 0, the network should be fine-tuned. When the score is larger than $m$, positive and negative samples will extracted from this frame for the next fine-tuning. When the network is being updated, it will iterated 10 times and will stop iteration if the loss is less than $l$, which is 0.01. We could see the results of frame 2-9 in Fig.~\ref{fig:fc_i} and know some parameters from Table.~\ref{fc_t}. In Table.~\ref{fc_t}, scores in frame 6-9 are less than the threshold $m$, so the network has to be updated. But the loss is very small in frame 6. The small loss means that the parameters in the network is been updated very well. In frame 7,8 and 9, we have to use the same positive and negative samples from frame 1,2,4 and 5(whose scores are larger than 0) to update the network, and because the loss is very small, these updates are ineffective and time consuming. So we use the loss threshold $l$ to control the update and it will stop the fine-tuning when the loss is less than $l$.\\
We could see from the Fig.~\ref{fig:fc_i} that it can track the object very well even thought the scores in frame 6,7,8 and 9 are less than the threshold $m$. It means that $m$ is not suitable for this sequence. But there are 100 sequences in OTB100 and we can not change the threshold in each sequence. So we could just use the loss threshold $l$ to control the fine-tuning of the network.\\
The last row of Table.~\ref{fc_t} is the numbers of iteration which the network needs. We can see that it only needs $7+6+1+1+1=16$ iterations in these frames. But in MDNet, the network always has to be updated 10 times and the total number of iteration is $10\times5=50$. We know that updating the network when tracking will have huge influence on the speed of the tracker. By this method, we can reduce a large amount of time and it will not affect the tracking result. Because the iterations we delete are useless and ineffective. 

\begin{table}
	\centering
	\caption{the states of fine-tuning in the frame 2-9 of Ironman}
	\begin{tabular}{ccccccccc}   
		\toprule
		frame        & 2     & 3      & 4     & 5     & 6      & 7      & 8      & 9      \\  
		\midrule        
		score        & 0.392 & -1.443 & 3.199 & 1.444 & -5.294 & -3.340 & -2.419 & -1.221 \\ 
		loss         & -     & 0.03   & -     & -     &  0.006 & 0.004  & 0.003  & 0.001   \\ 
	    iteration    & -     & 7      & -     & -     &  6     & 1      & 1      & 1      \\ 
		\bottomrule  
		\label{fc_t}
	\end{tabular}
\end{table}

Moreover, hard negative mining~\cite{32} and bounding box regression~\cite{25} have been employed. Experiments~\cite{26} show that both of them can improve the tracking performance. Hard negative mining is a trick to select effective samples. The negative samples will be sent to the network to get scores. And the samples with higher scores are taken as the hard negative samples. We use these hard negative samples instead of all the negative samples to update the fully-connected layers. The bounding box regression is just like what they used in R-CNN~\cite{11}. A simple linear regression model is trained to predict the target region using the features of the $3\textsuperscript{rd}$ convolution layer. The training of the bounding box regression is only conducted with the first frame. The learning rate is set to 0.0005 in the first frame and it is set to 0.0015 for online updating.\\
\section{Experiment}
We evaluated our network on two datasets: object tracking benchmark (OTB) and VOT2016. Our code is implemented in MATLAB with matconvnet~\cite{35}. It runs at about 7 fps on Tesla p40 GPU.
\subsection{Experiment on OTB}
\begin{figure} 	
	\centering 	
		\includegraphics[width=0.24\linewidth]{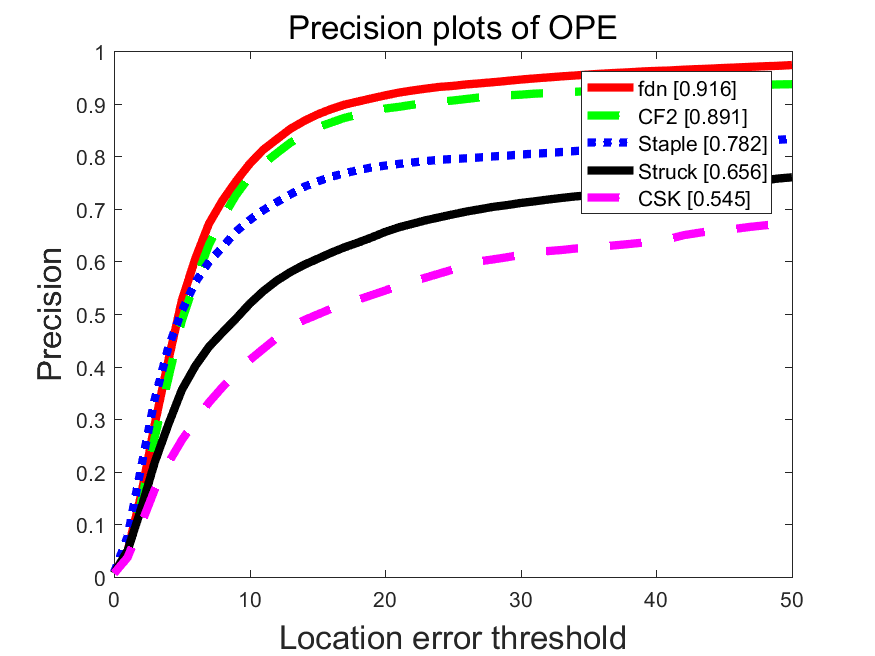}
		\includegraphics[width=0.24\linewidth]{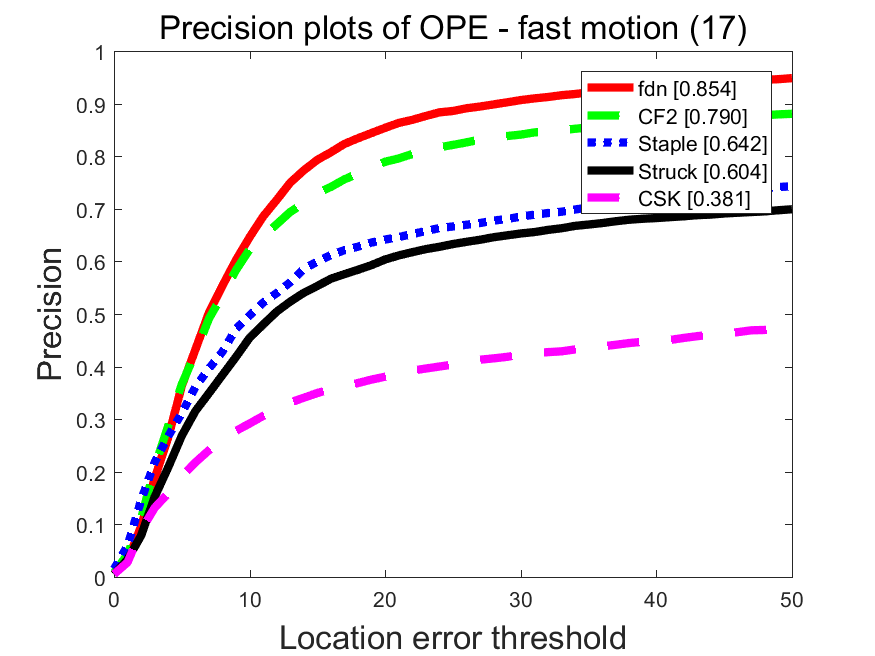}
		\includegraphics[width=0.24\linewidth]{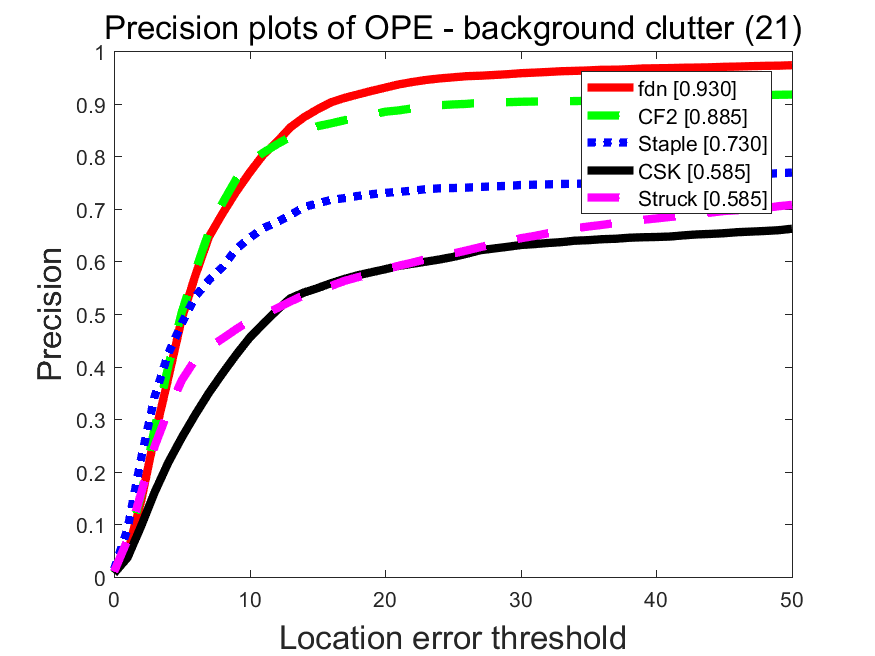}
		\includegraphics[width=0.24\linewidth]{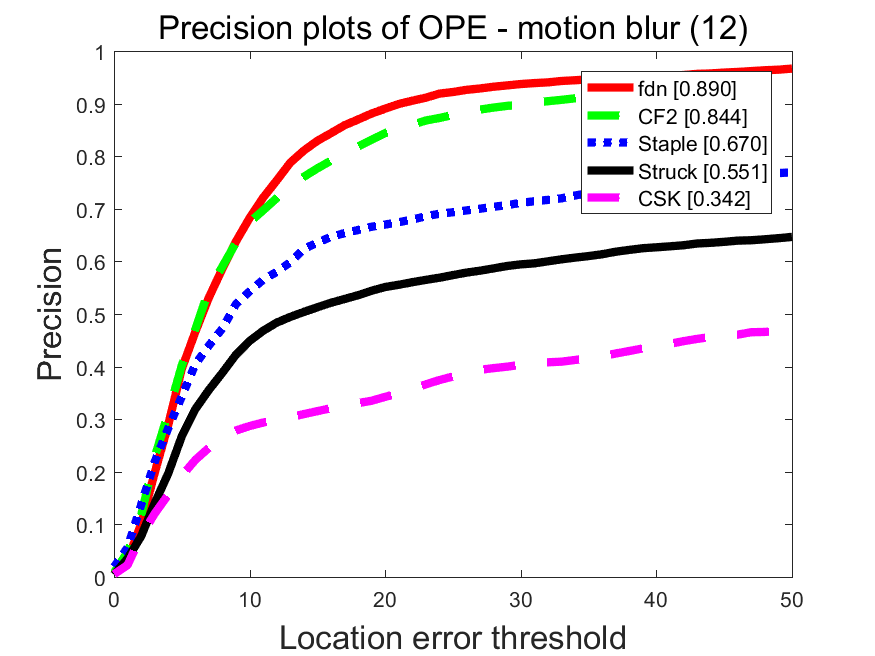}   

		\includegraphics[width=0.24\linewidth]{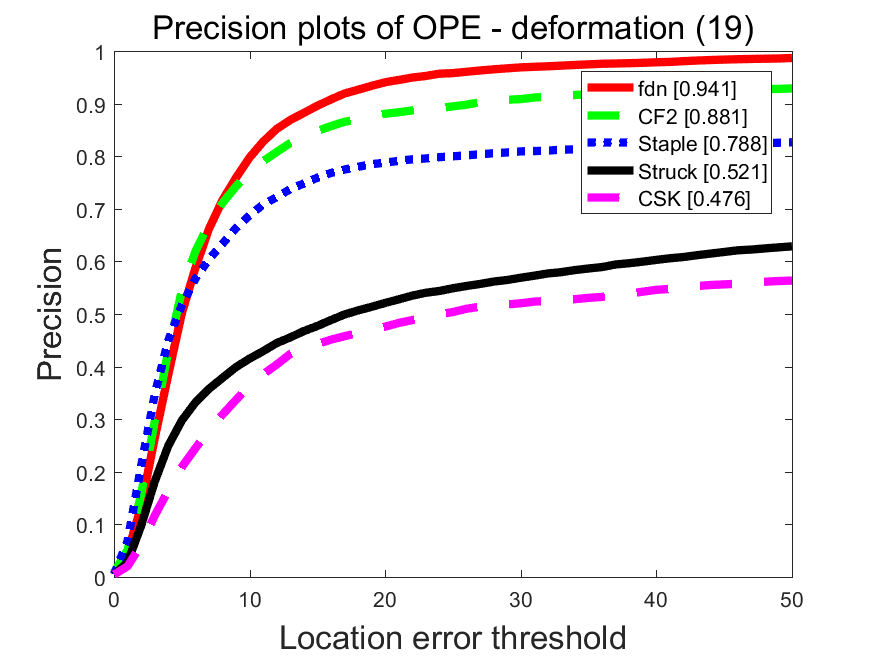}
		\includegraphics[width=0.24\linewidth]{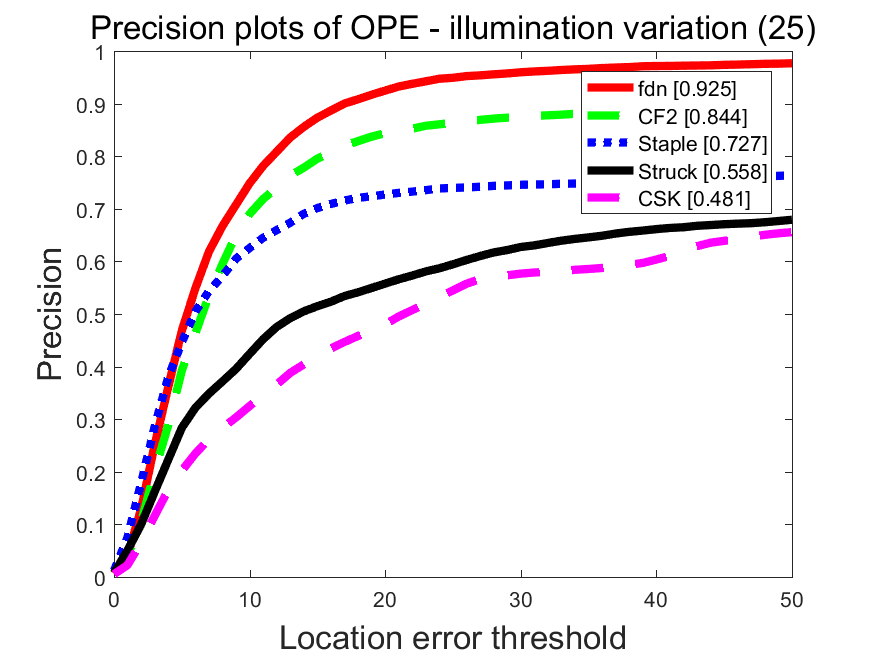}
		\includegraphics[width=0.24\linewidth]{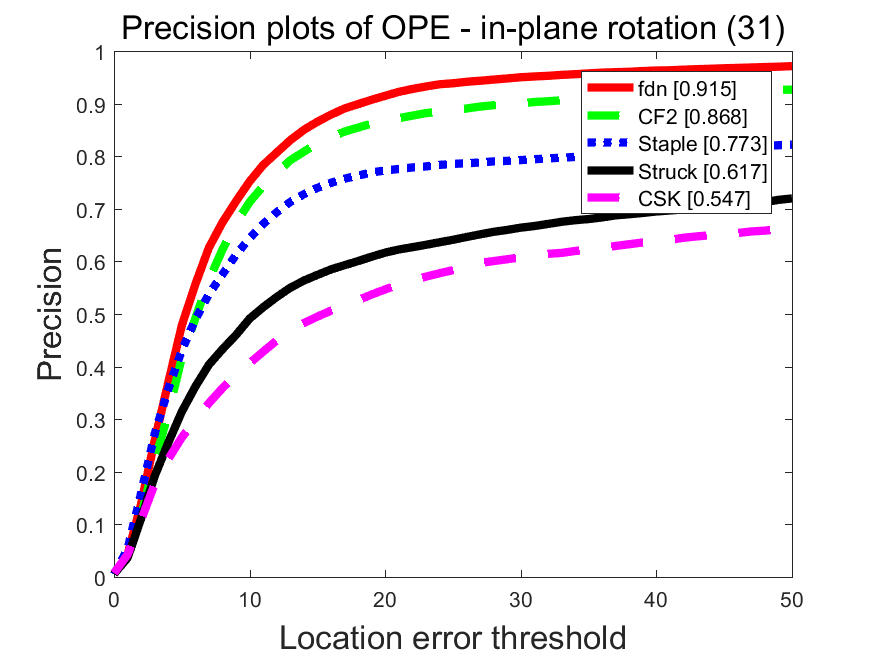}
		\includegraphics[width=0.24\linewidth]{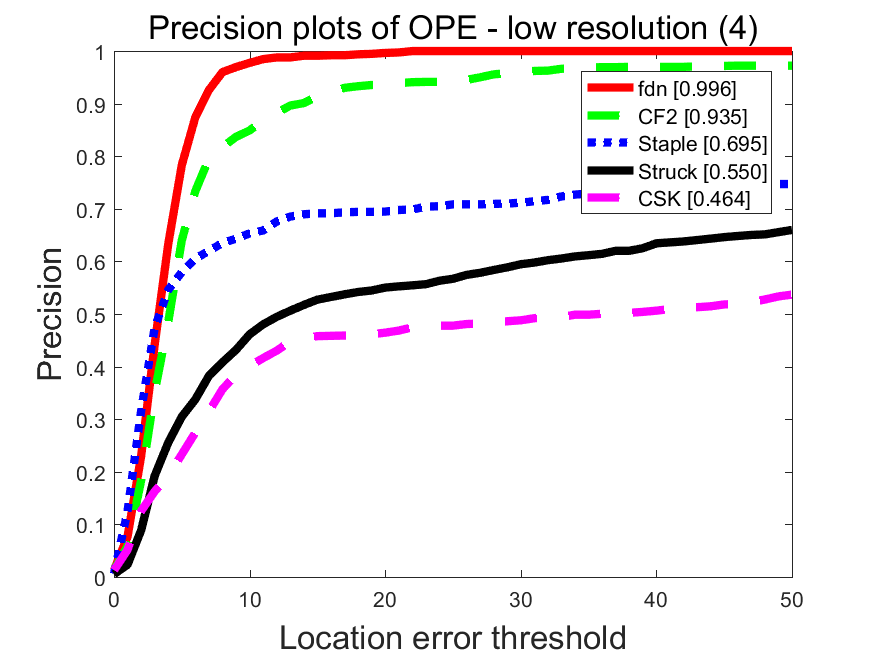}  
	
		\includegraphics[width=0.24\linewidth]{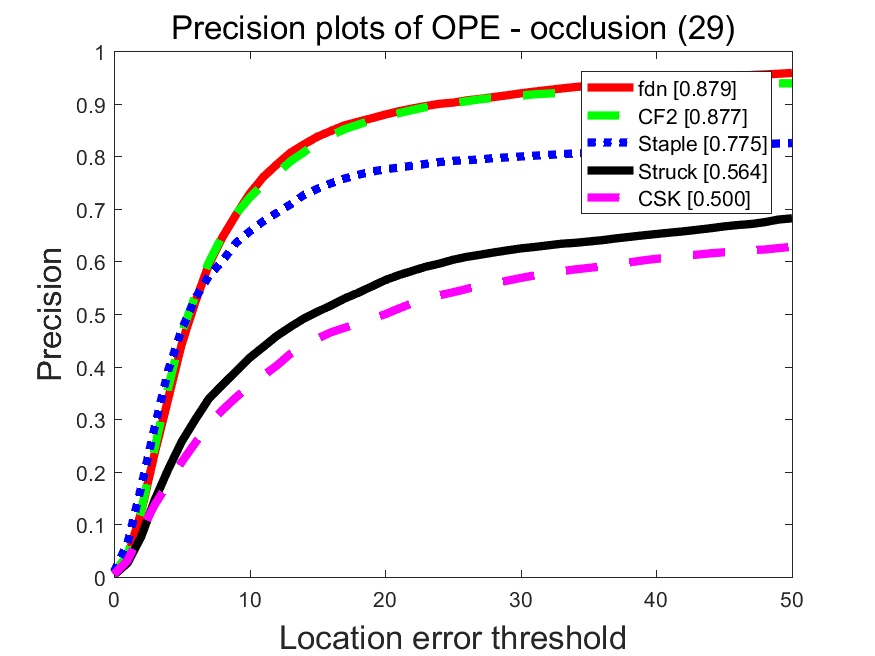}
		\includegraphics[width=0.24\linewidth]{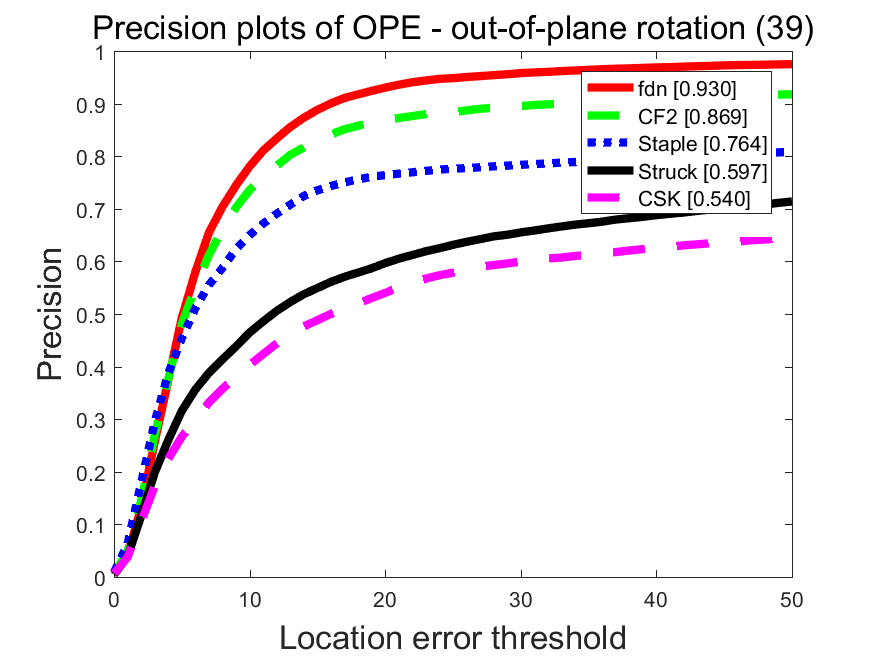}
		\includegraphics[width=0.24\linewidth]{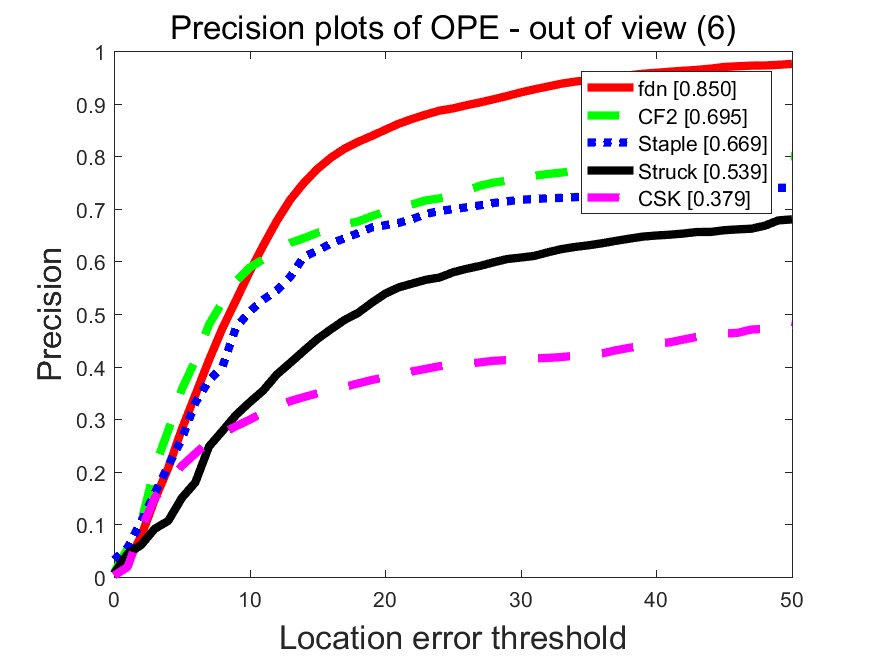}
		\includegraphics[width=0.24\linewidth]{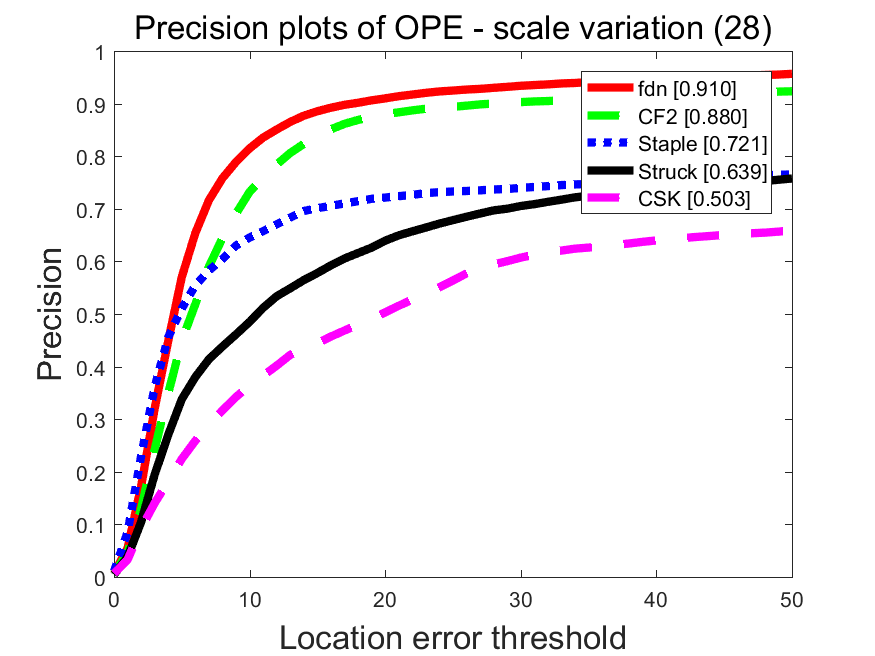}   
	
	\caption{Results on OTB2013}
	\label{fig:OTB2013}	
\end{figure}

\begin{figure} 	
	\centering 	
	\includegraphics[width=0.24\linewidth]{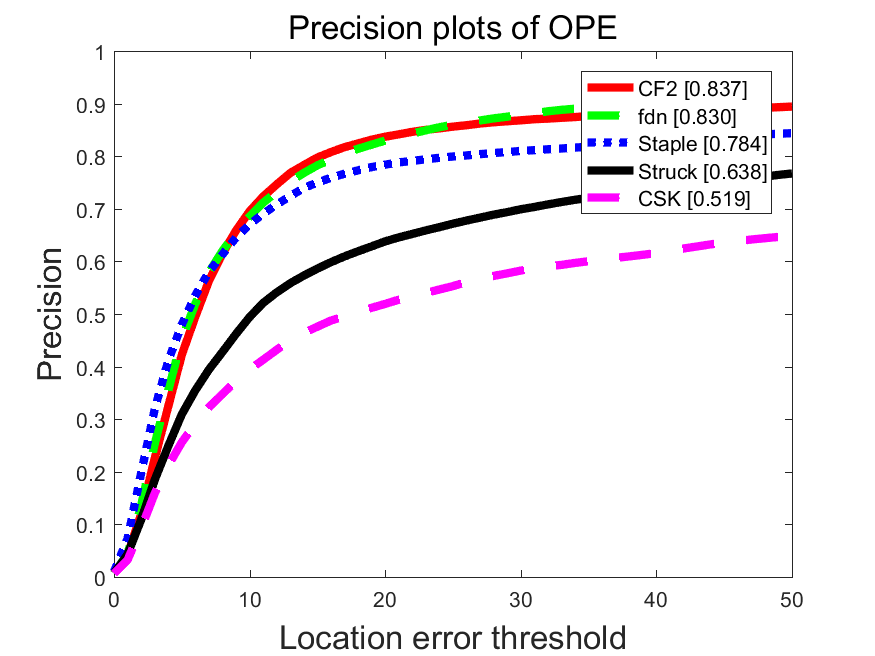}
	\includegraphics[width=0.24\linewidth]{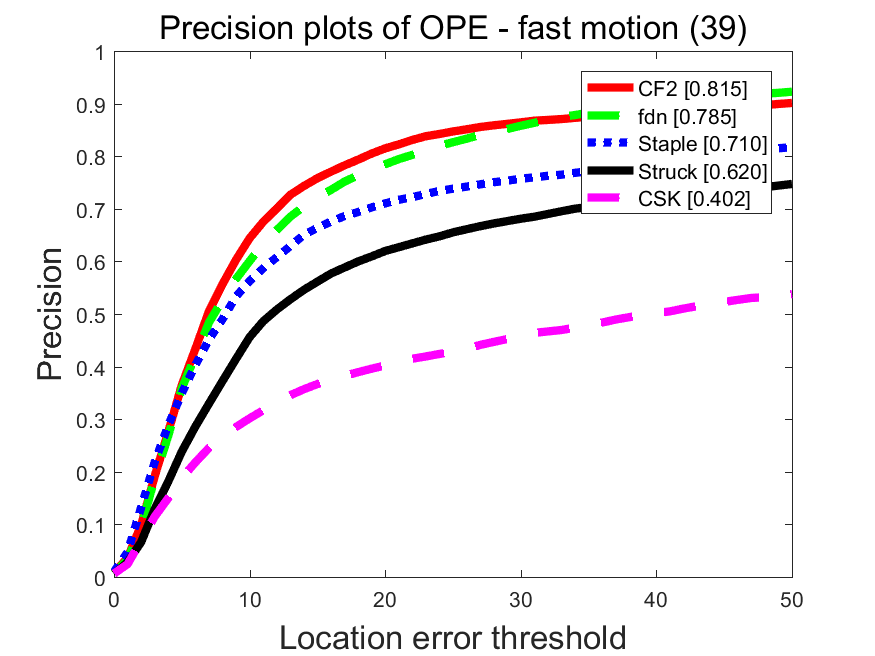}
	\includegraphics[width=0.24\linewidth]{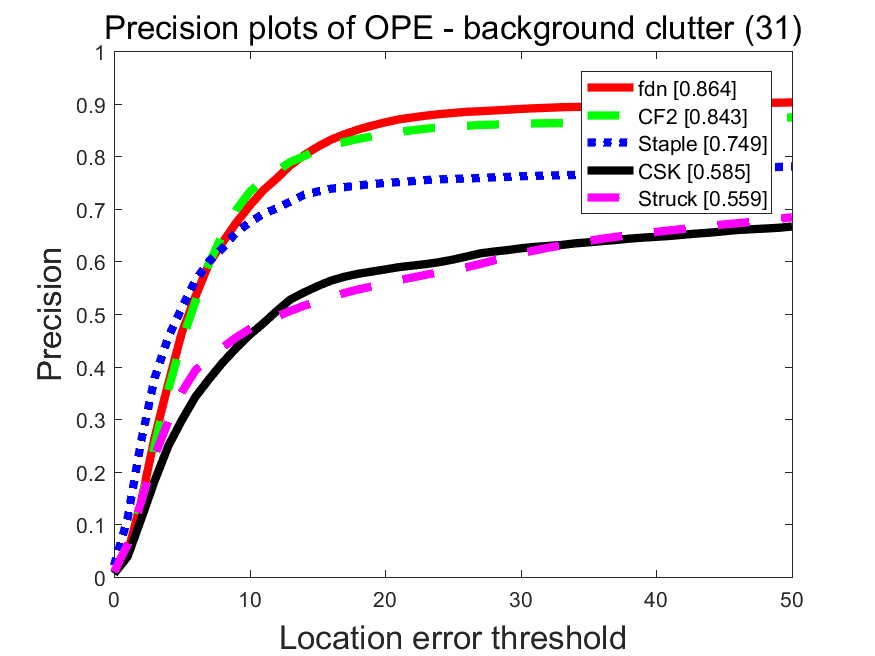}
	\includegraphics[width=0.24\linewidth]{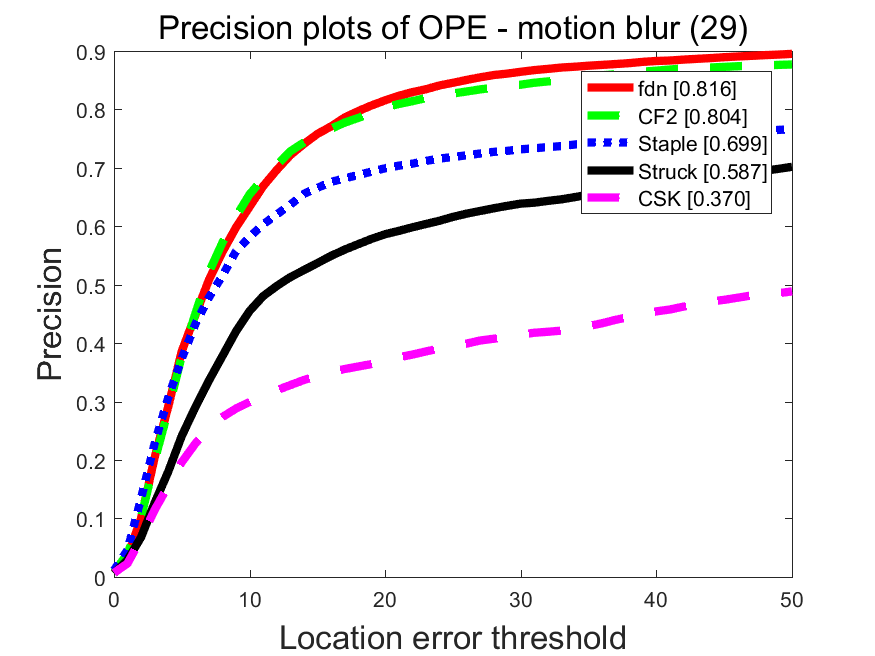}   

	\includegraphics[width=0.24\linewidth]{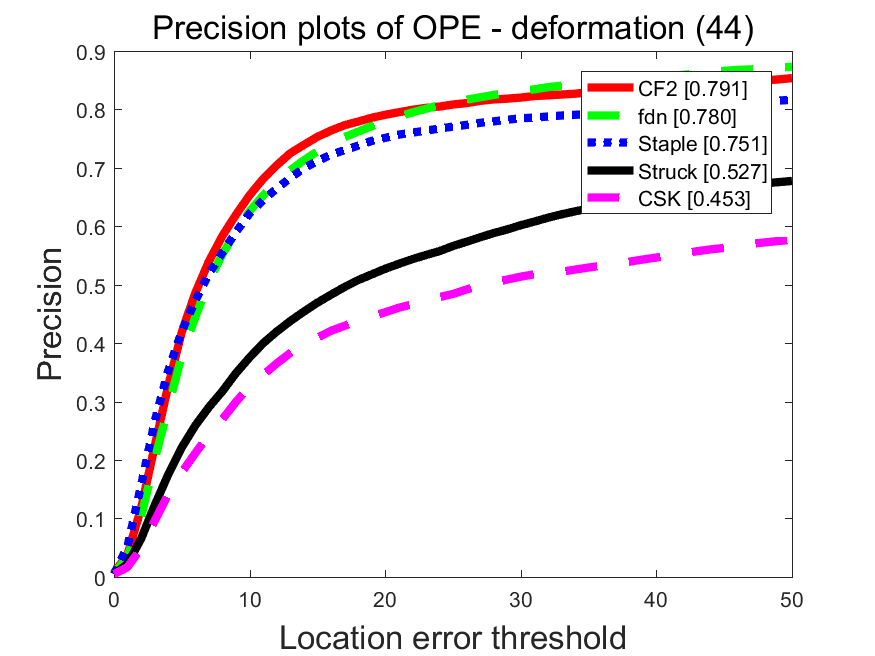}
	\includegraphics[width=0.24\linewidth]{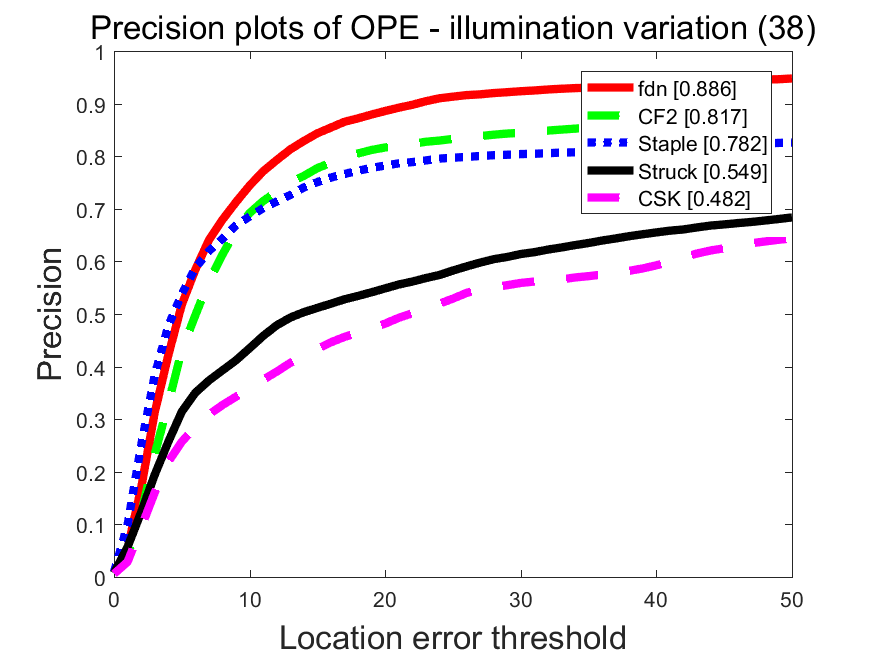}
	\includegraphics[width=0.24\linewidth]{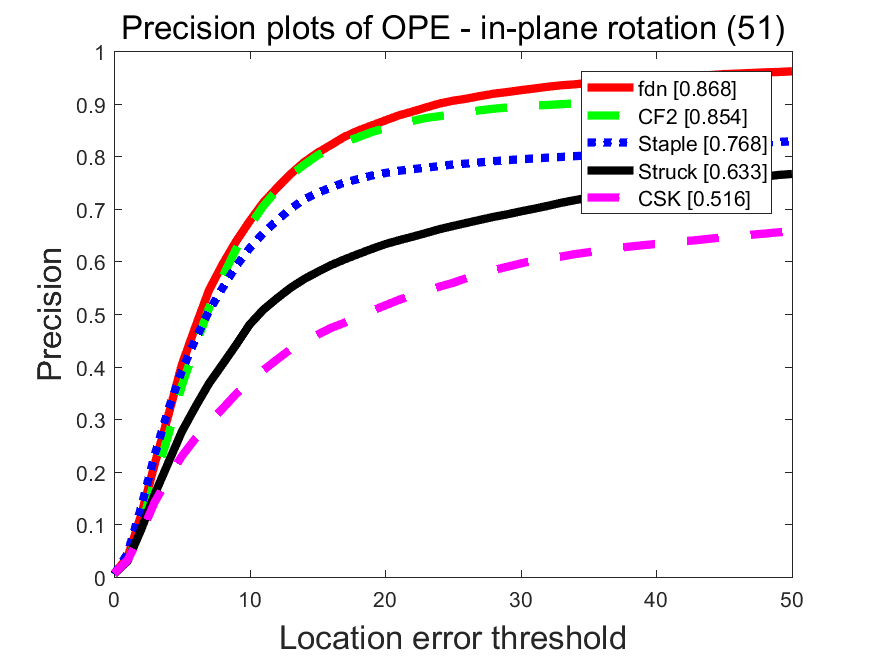}
	\includegraphics[width=0.24\linewidth]{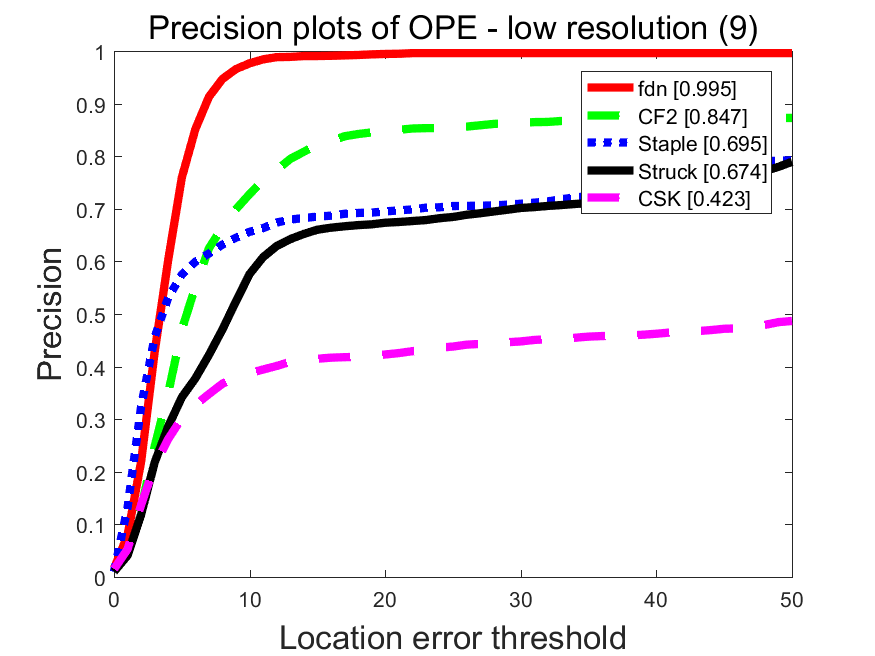}  
	
	\includegraphics[width=0.24\linewidth]{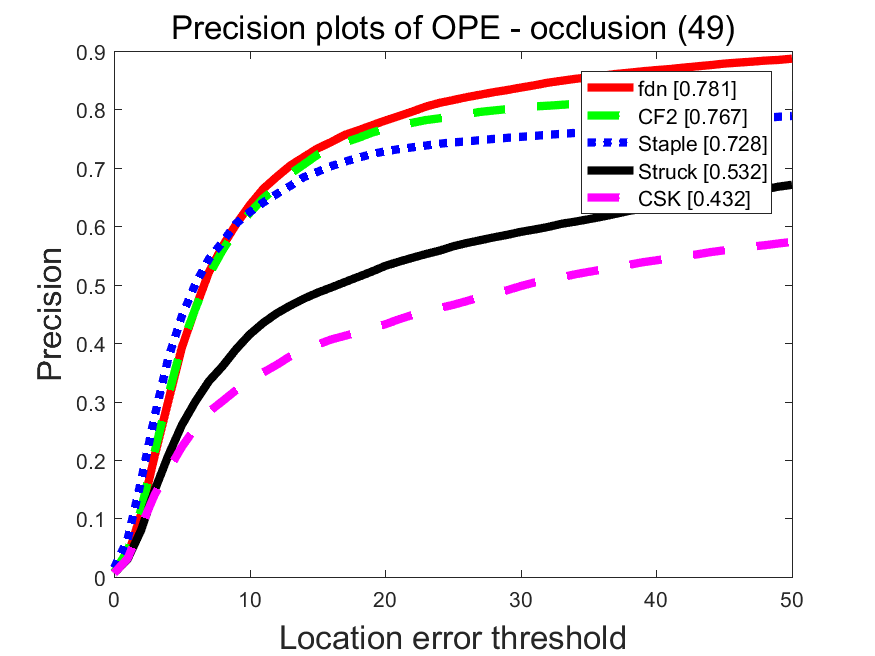}
	\includegraphics[width=0.24\linewidth]{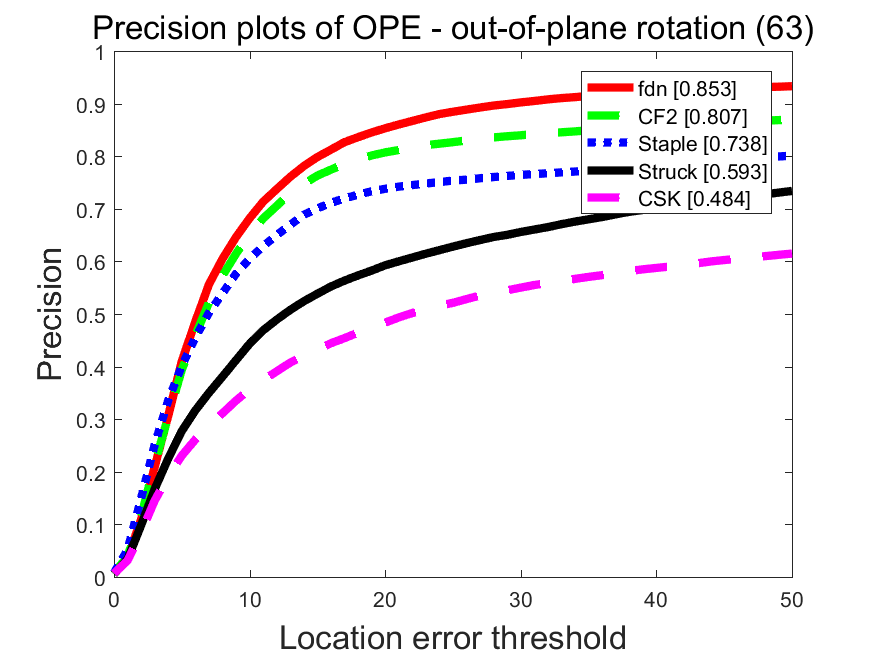}
	\includegraphics[width=0.24\linewidth]{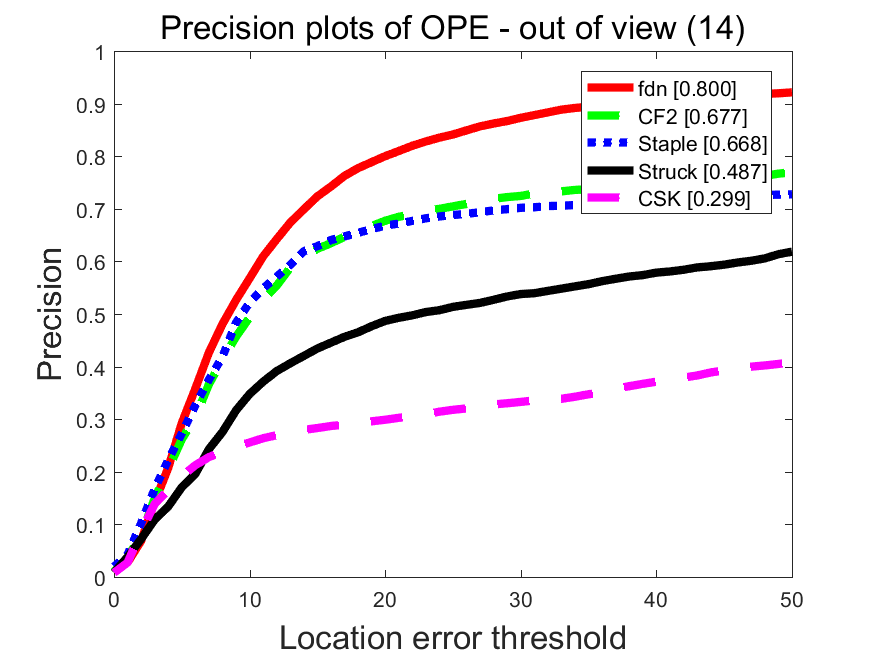}
	\includegraphics[width=0.24\linewidth]{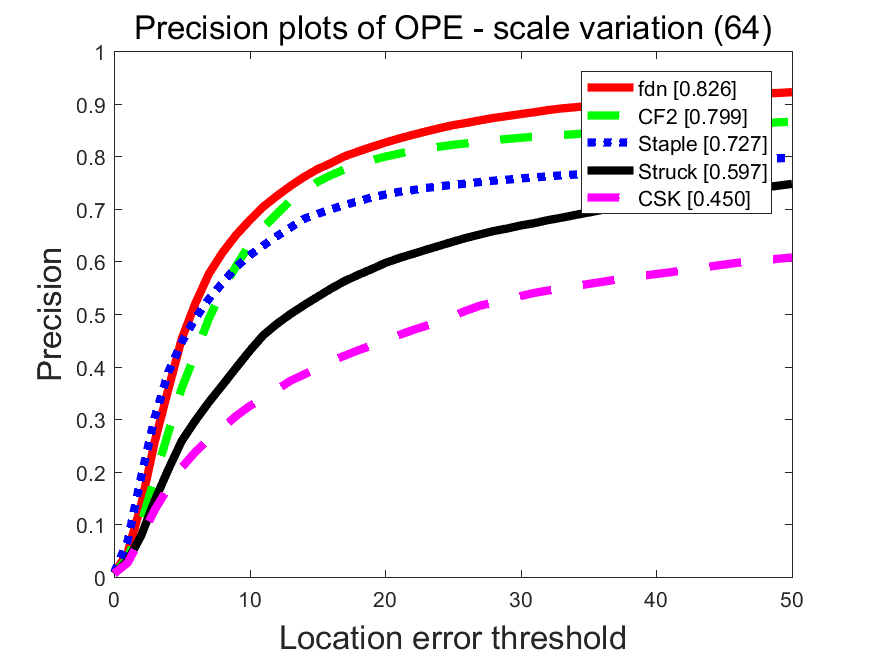}   
	
	
	\caption{Results on OTB100}
	\label{fig:OTB100}	
\end{figure}

\begin{figure} 	
	\centering 	
		
	\includegraphics[width=0.24\linewidth]{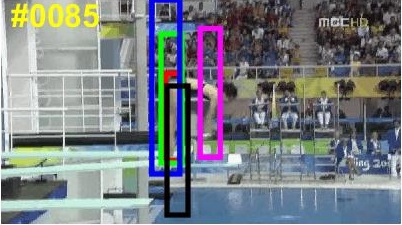}
	\includegraphics[width=0.24\linewidth]{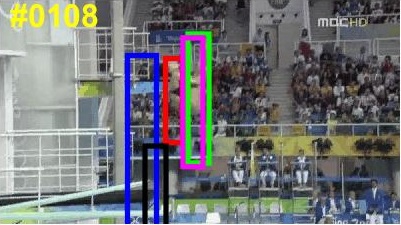}
	\includegraphics[width=0.24\linewidth]{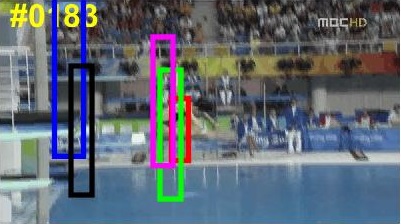}
	\includegraphics[width=0.24\linewidth]{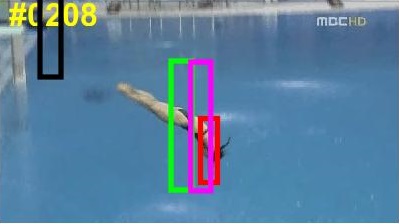}

	\includegraphics[width=0.24\linewidth]{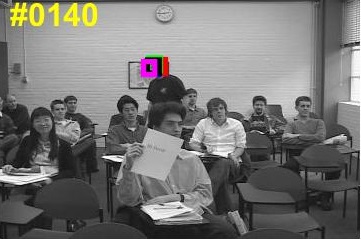}
	\includegraphics[width=0.24\linewidth]{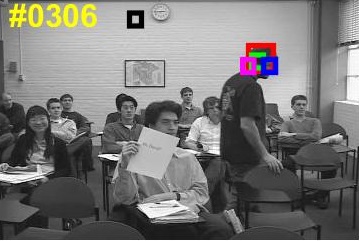}
	\includegraphics[width=0.24\linewidth]{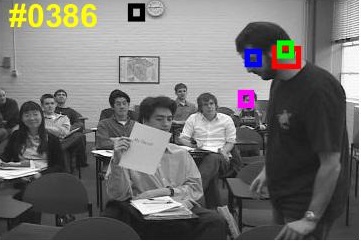}
	\includegraphics[width=0.24\linewidth]{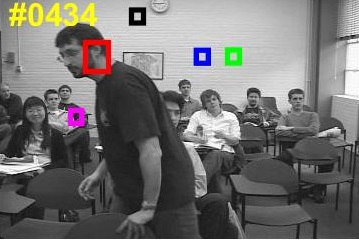}

	\includegraphics[width=0.24\linewidth]{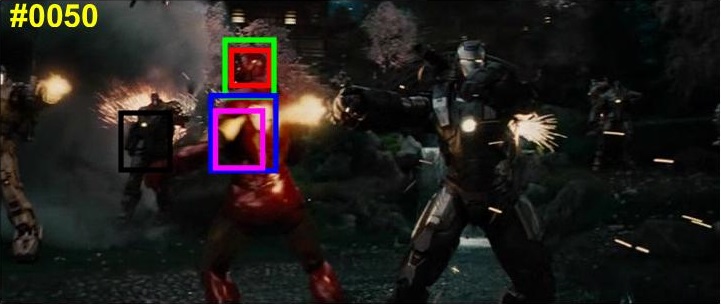}
	\includegraphics[width=0.24\linewidth]{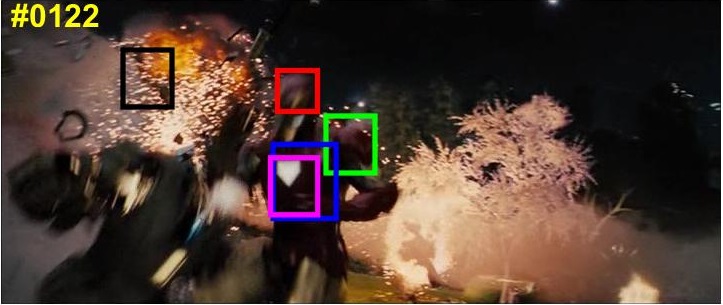}
	\includegraphics[width=0.24\linewidth]{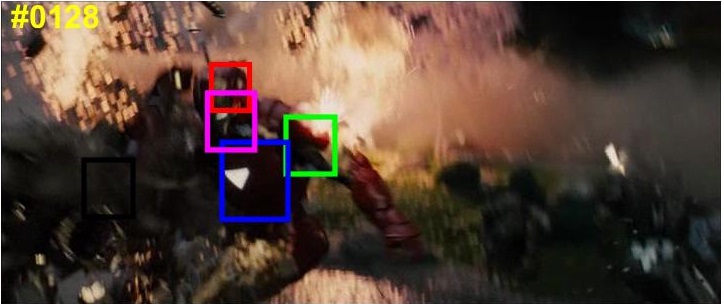}
	\includegraphics[width=0.24\linewidth]{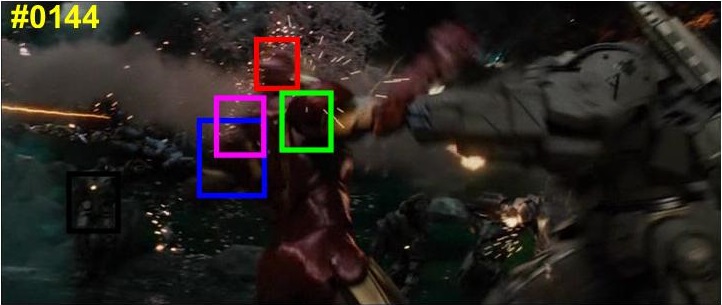}   
	
	\includegraphics[width=0.24\linewidth]{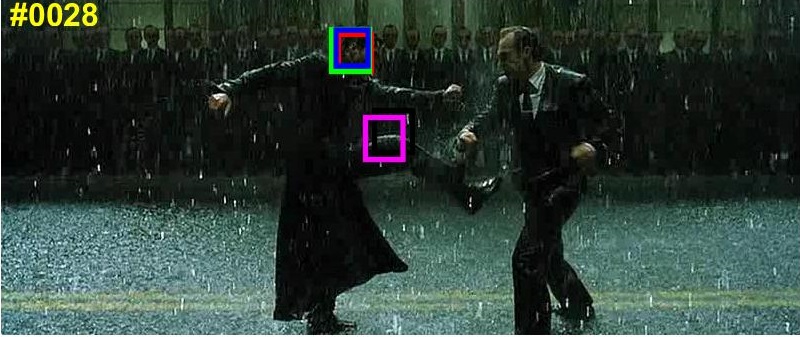}
	\includegraphics[width=0.24\linewidth]{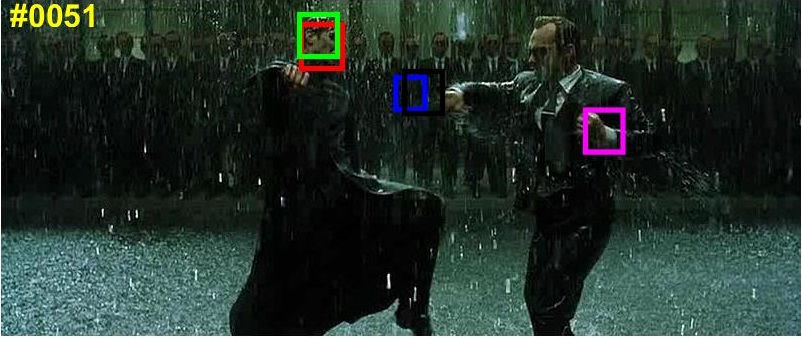}
	\includegraphics[width=0.24\linewidth]{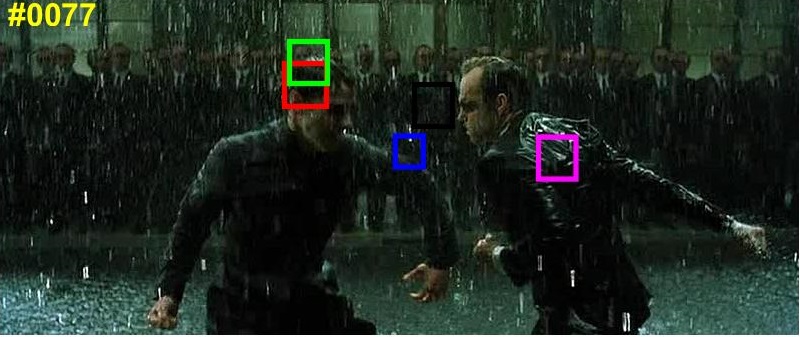}
	\includegraphics[width=0.24\linewidth]{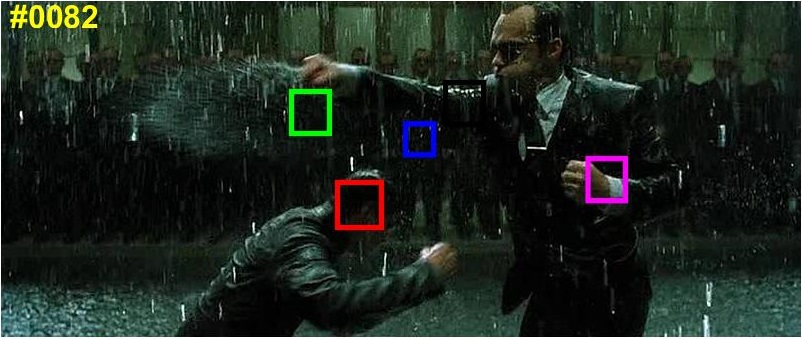}  
	
	\includegraphics[width=0.24\linewidth]{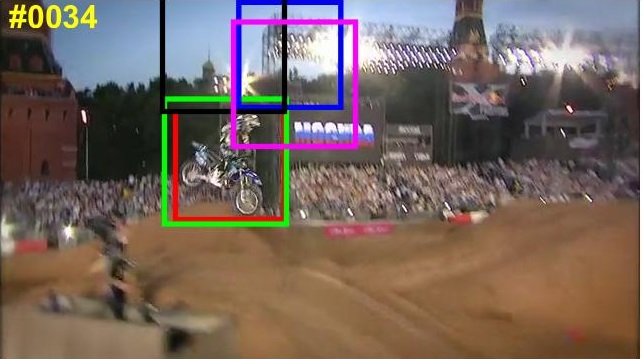}
	\includegraphics[width=0.24\linewidth]{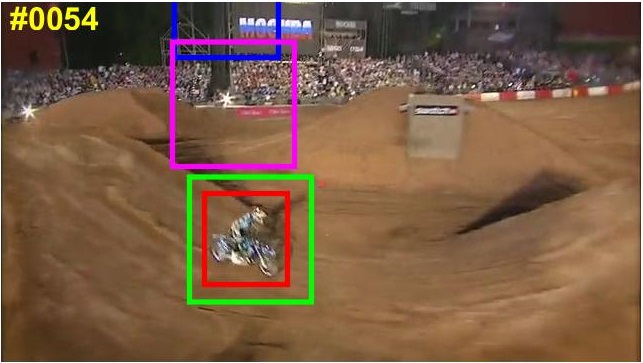}
	\includegraphics[width=0.24\linewidth]{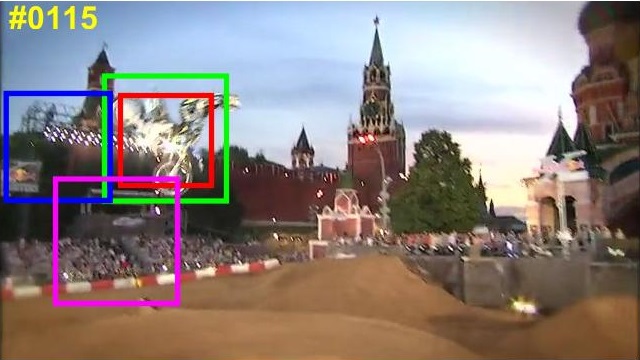}
	\includegraphics[width=0.24\linewidth]{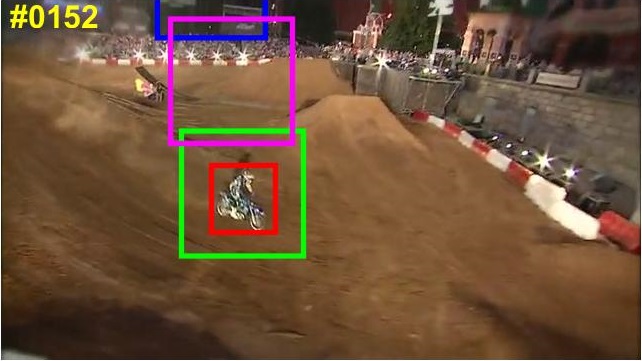}   
	
	\includegraphics[width=0.80\linewidth]{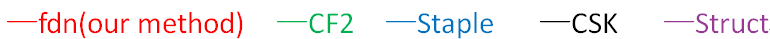}   
	
	\caption{Comparison of some visual tracking examples}
	\label{fig:OTBresults}	
\end{figure}

OTB~\cite{39} is a popular tracking benchmark. Each of those vedios is labeled with attributes. The evaluation is based on two metrics: center location and bounding box overlap ratio. Three evaluation methods have been employed: OPE(one pass evaluation), TRE(temporal robustness evaluation) and SRE(spatial robustness evaluation). The proposed method has been compared with state of the art methods including CSK~\cite{25}, Struct~\cite{12}, CF2~\cite{23} and Staple~\cite{2}. To train the network, 58 videos from VOT2013~\cite{47}, VOT2014~\cite{48} and VOT2015~\cite{49} have been employed, excluding the videos included in OTB100. The results of OTB2013 and OTB100 are shown in Fig.~\ref{fig:OTB2013}. and Fig.~\ref{fig:OTB100}., respectively. And some examples of the visual tracking results are shown in Fig.~\ref{fig:OTBresults}. From these results it could be concluded that superior or comparable performance has been obtained via the proposed method.\\
The comparison of MDNet and our network is in Table.~\ref{mdnet_otb2013}. Our method is $7\times$ faster than MDNet with only a small drop in precision.

\subsection{Experiment on VOT2016}
\begin{figure}
	\centering
	\subfigure[VOT2016 accuracy-robustness plot. The better trackers are closer to the top-right corner.]{
		\begin{minipage}[b]{0.53\textwidth}
			\includegraphics[width=1\textwidth]{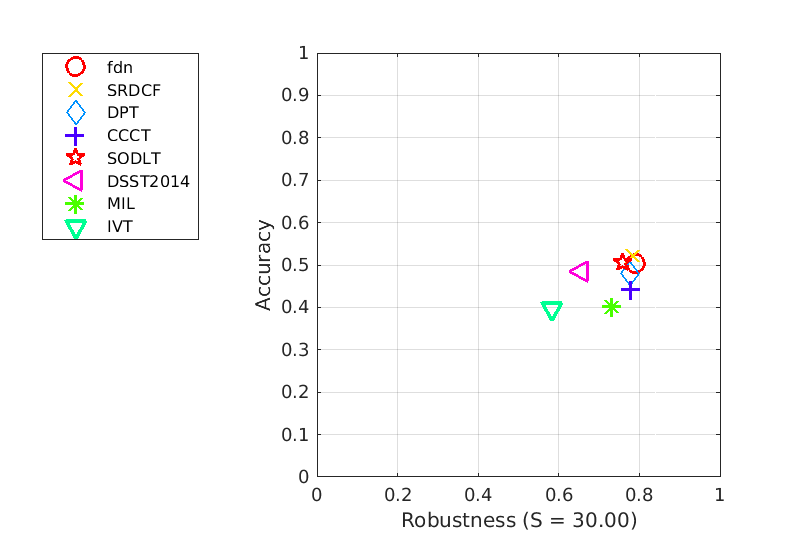}
		\end{minipage}
	}
	\subfigure[VOT2016 ranking in terms of expected average overlap]{
		\begin{minipage}[b]{0.39\textwidth}
			\includegraphics[width=1\textwidth]{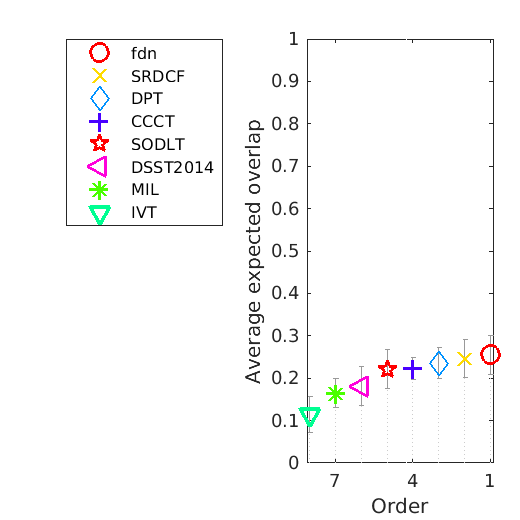}
		\end{minipage}
	}
	\caption{Results on VOT2016} \label{fig:VOTresults}
\end{figure}

For further comparison, the algorithm has also been evaluated on the VOT2016~\cite{50} which consists of 60 videos. VOT is a challenging competition organized for several years. The benchmark will re-initialize the targets when the tracker fails and evaluate module reports with both accuracy and robustness in terms of the bounding box overlap ratio and the number of failures, respectively. We compare our tracker with SRDCF~\cite{51},DPT~\cite{52},CCCT~\cite{53},DSST2014~\cite{54},SO-DLT~\cite{55}, MIL~\cite{56} and IVT~\cite{57}. 84 videos from OTB100 are used to train the network, which do not include the sequences in VOT2016. The results are shown in Fig.~\ref{fig:VOTresults}. As illustrated in Fig.~\ref{fig:VOTresults}    , our tracker shows great results both in accuracy and robustness. It also shows that our method rank $1\textsuperscript{st}$ in these trackers according to EAO(Expected Average Overlap) criterion. 

\subsection{Comparisons of RoIAlign and RoIPool}
\begin{figure} 	
	\centering 	
	
	\includegraphics[width=0.45\linewidth]{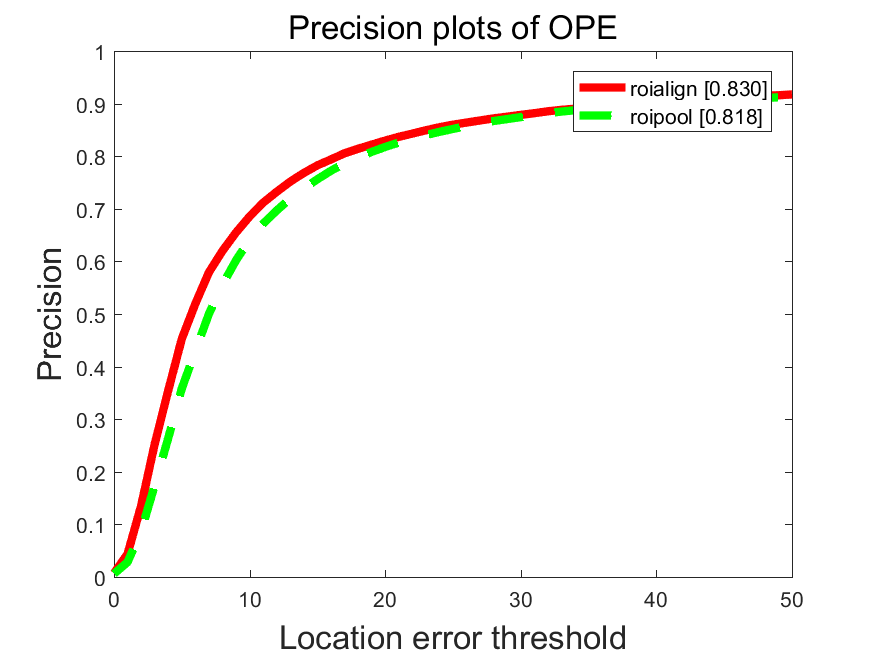}
	\includegraphics[width=0.45\linewidth]{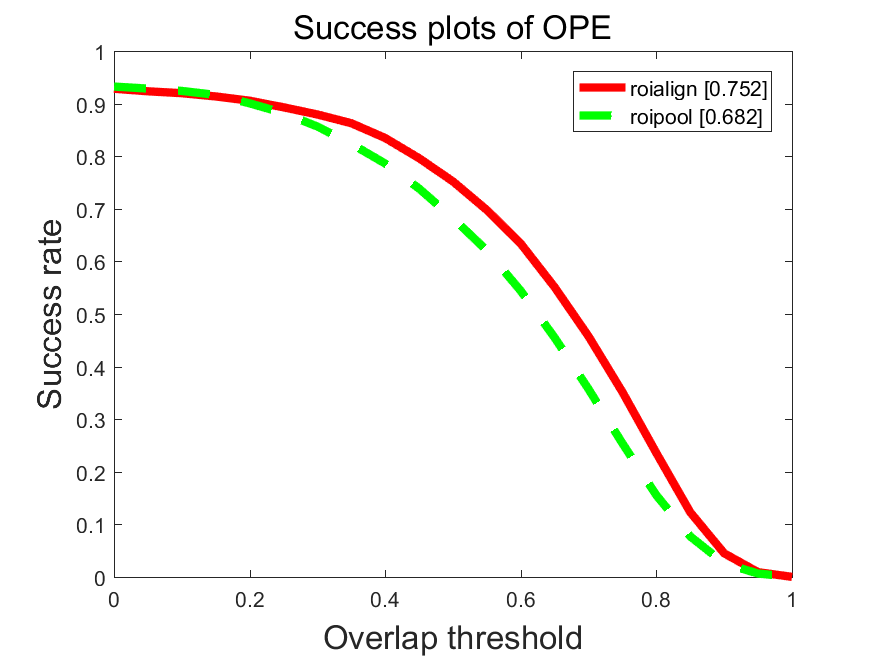}	
	\caption{The results of RoIPool and RoIAlign on OTB100}
	\label{fig:roipool}	
\end{figure}

Most of the networks using RoIPool from fast-RCNN to accelerate the convolution have obtained good performance on detection and classification task. However, for tracking tasks, the localization error will continue to be accumulated and enlarged along the working of the tracker. Even a small error of the current frame will affect the ones after it. So we use RoIAlign instead of RoIPool to improve the target localization precision. Fig.~\ref{fig:roipool}. shows the results of RoIAlign and RoIPool on OTB100.
\subsection{Influence of the Number of Convolution or Fully-connected Layers}
\begin{figure} 	
	\centering 	
	\includegraphics[width=0.45\linewidth]{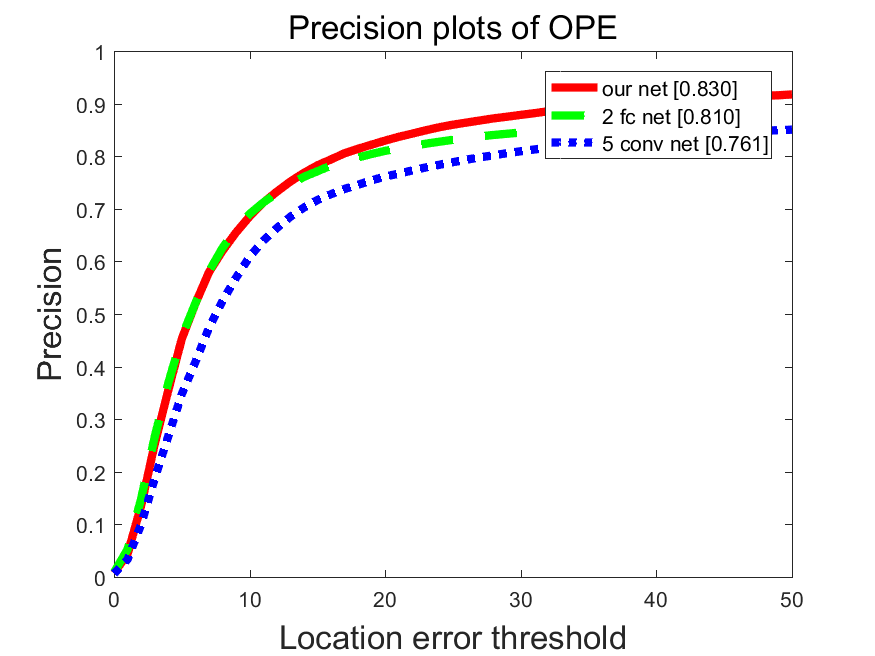}
	\includegraphics[width=0.45\linewidth]{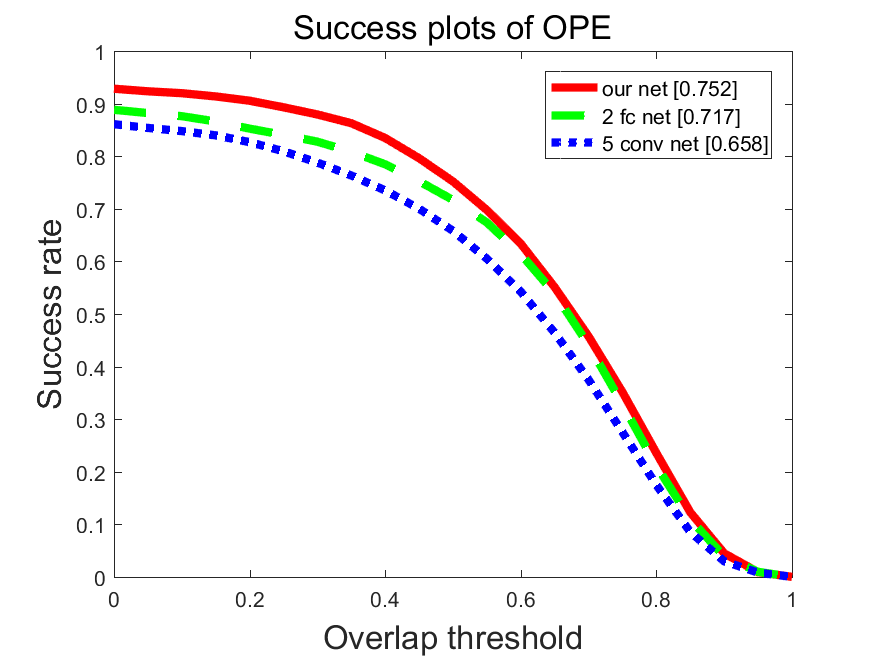}	
	\caption{The results of our network,5 convolution network and 2 fully-connected network on OTB100}
	\label{fig:netchange}	
\end{figure}

Considering we just use 3 convolution layers of the VGG-M network, and it has 5 convolution layers in the VGG-M network. More convolution layers may lead to better performance, so we use 5 convolution layers to build a network, and the rest part are kept the same with our previous method. The comparison resuts on OTB100 are given in Fig. 8. In addition, considering there are only two categories, there is no need to use more parameters in the fully-connected layers. Therefore, to evaluate the influence of the number of fully-connected layers, a network with just two fully-connected layers is constructed and the rest are the same. The results on OTB100 are also shown in Fig.~\ref{fig:netchange}. We can see that the network structure used in the proposed method obtains the best performance.\\
More convolution layers may lead to a better results in detection. But a deep CNN is less effective for precise target localization since the spatial information tends to be diluted as a network goes deeper~\cite{58}. Localization information is very important in tracking. So it explain why more convolution may not result in a better precision.

\begin{table}
	\centering
	\caption{comparison of MDNet and our method on OTB2013}
	\begin{tabular}{cccc}   
		\toprule
		             & precision & success rate & speed  \\
		\midrule       
		MDNet        &0.948      & 0.708        &  1 fps   \\
		Our Method   &0.916      & 0.667        &  7 fps      \\ 
		
		\bottomrule  
		\label{mdnet_otb2013}
	\end{tabular}
\end{table}

\subsection{Do We Need to Train the Network with Videos}
\begin{figure} 	
	\centering 	
	\includegraphics[width=0.45\linewidth]{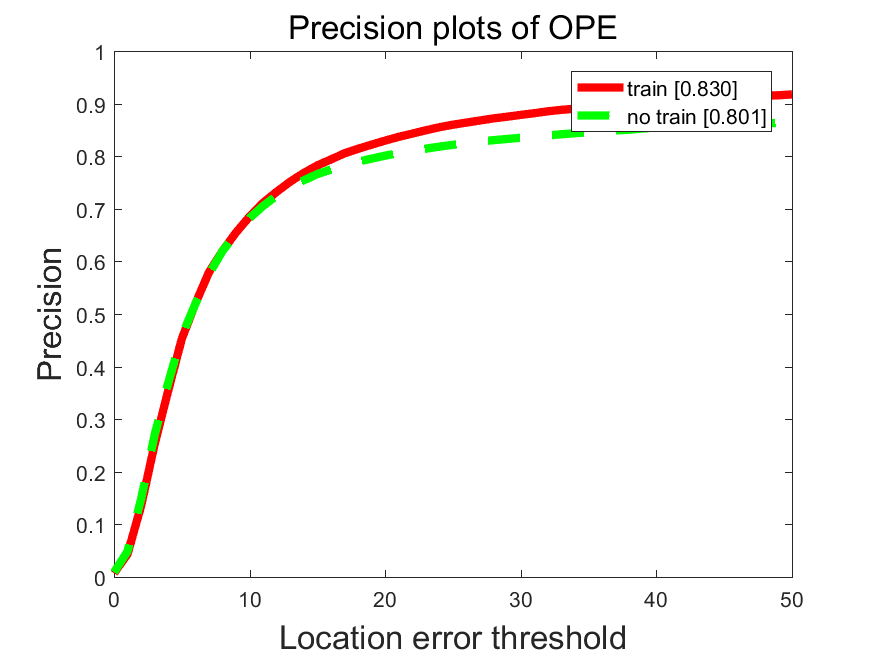}
	\includegraphics[width=0.45\linewidth]{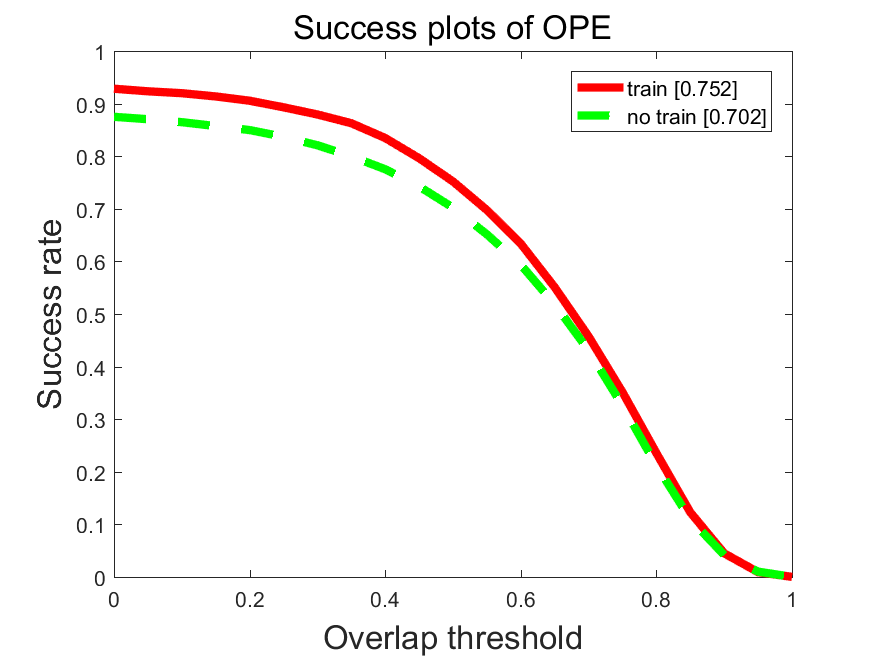}	
	\caption{The results of train and no train network on OTB100}
	\label{fig:notrain}	
\end{figure}

We all know that convolution layers are in charge of extracting features and fully-connected layers are the classifiers to distinguish objects. The convolution layers in our network are trained on ImageNet and it is a good feature extractor. Maybe it does not need to be trained again and fine-tuning the fully-connected layers can accomplish tracking. So we just use first 3 convolution layers of VGG-M and put three randomly initialized fully-connected layers after them to build a network. The results on OTB100 are shown in Fig.~\ref{fig:notrain}. It demonstrates that the network still can work but not as well as the trained one.

\section{Conclusions}
We propose a fast dynamic convolution neural network for visual tracking. There are 3 convolution layers, 3 fully-connected layers and a RoIAlign layer between them. The RoIAlign layer will accelerate the convolution and performs better than RoIPool with higher target localization precision. The last fully-connected layer has $k$ branches which correspond to $k$ videos during training. When tracking, the last fully-connected layer is deleted and a new one which is randomly initialized is added. The first frame is used to train the fully-connected layers and update the parameters. A new strategy of fine-tuning the fully-connected layers is used to accelerate the update when tracking. Finally about 7 fps on GPU is reached and  outstanding performance on two tracking benchmarks, OTB and VOT2016, has been obtained.

%

\bibliography{acml18}






\end{document}